\documentclass{article}

\usepackage{arxiv}

\usepackage[utf8]{inputenc} % allow utf-8 input
\usepackage[T1]{fontenc}    % use 8-bit T1 fonts
\usepackage{hyperref}       % hyperlinks
\usepackage{url}            % simple URL typesetting
\usepackage{booktabs}       % professional-quality tables
\usepackage{amsfonts}       % blackboard math symbols
\usepackage{nicefrac}       % compact symbols for 1/2, etc.
\usepackage{microtype}      % microtypography
\usepackage{lipsum}		% Can be removed after putting your text content
\usepackage{graphicx}
\usepackage[square, comma, sort&compress, numbers]{natbib}
\usepackage{doi}
\usepackage{amsmath}
\usepackage{subcaption}
\usepackage{multirow}
\usepackage{algorithm}
\usepackage{algorithmicx}
\usepackage{algpseudocode}
\usepackage{tabularx}
\usepackage[flushleft]{threeparttable}
\usepackage{lineno}

\newcommand{\return}[1]{\textbf{\color{blue} return} #1}
\algnewcommand{\LineComment}[1]{\State \textit{\color{orange} \# #1}}

\newcommand{\set}[1]{\{#1\}}

\usepackage[colorinlistoftodos,prependcaption,textsize=tiny]{todonotes}

\newcommand{\subtext}[1]{$_{\text{#1}}$}

\title{GEM-2: Next Generation Molecular Property Prediction Network by Modeling Full-range Many-body Interactions}

\author{ 
    Lihang Liu\thanks{Equal contribution} \\
	Baidu Inc. \\
	\texttt{liulihang@baidu.com} \\
	\And
	Donglong He$^*$ \\
	Baidu Inc. \\
	\texttt{hedonglong@baidu.com}
	\And
	Xiaomin Fang\thanks{Corresponding authors} \\
	Baidu Inc. \\
	\texttt{fangxiaomin01@baidu.com} \\
	\And
	Shanzhuo Zhang \\
	Baidu Inc. \\
	\texttt{zhangshanzhuo@baidu.com}
	\And
	Fan Wang\footnotemark[2] \\
	Baidu Inc. \\
	\texttt{wang.fan@baidu.com}
	\And
	Jingzhou He \\
	Baidu Inc. \\
	\texttt{hejingzhou@baidu.com}
	\And
	Hua Wu \\
	Baidu Inc. \\
	\texttt{wu\_hua@baidu.com}
}

\date{}

\renewcommand{\shorttitle}{GEM-2}

\begin{document}
\maketitle

\begin{abstract}
% \old{Physically, the properties of a molecule are determined by its own electronic structure, which can be exactly described by the Schrödinger equation. However, solving the Schrödinger equation for most molecules is extremely challenging due to long-range interactions in the behavior of a quantum many-body system. 
% While deep learning methods have proven to be effective in molecular property prediction, we design a novel method, namely GEM-2,}

Molecular property prediction is a fundamental task in the drug and material industries. Physically, the properties of a molecule are determined by its own electronic structure, which is a quantum many-body system and can be exactly described by the Schrödinger equation. Full-range many-body interactions between electrons have been proven effective in obtaining an accurate solution of the Schrödinger equation by classical computational chemistry methods, although modeling such interactions consumes an expensive computational cost. Meanwhile, deep learning methods have also demonstrated their competence in molecular property prediction tasks. Inspired by the classical computational chemistry methods, we design a novel method, namely GEM-2, which comprehensively considers full-range many-body interactions in molecules. Multiple tracks are utilized to model the full-range interactions between the many-bodies with different orders, and a novel axial attention mechanism is designed to approximate the full-range interaction modeling with much lower computational cost.
Extensive experiments demonstrate the overwhelming superiority of GEM-2 over multiple baseline methods in quantum chemistry and drug discovery tasks. The ablation studies also verify the effectiveness of the full-range many-body interactions.

\end{abstract}


\section{Introduction}
Molecular property prediction is a fundamental task in drug and material industries, evaluating the physical, chemical, and biological properties to assist the researchers in making decisions. In essence, the properties of a molecule are determined by its own electronic structure.
Theoretically, the electronic structure of molecules can be exactly described by the Schrödinger equation. A molecule with multiple electrons is a many-body system with quantum full-range interactions\cite{Defenu_2021}. Modeling the full-range many-body interactions have been proven critical in obtaining an accurate solution of the Schrödinger equation by classical computational chemistry methods. Two representative calculation methods to approximate the solution of the Schrödinger equation are density functional theory (DFT) and coupled-cluster theory (CC). The core idea of DFT is to represent all electrons with a single electron density functional, thus reducing the dimension of the problem\cite{Burke_Wagner_2013}. Previous work\cite{Peverati_Truhlar_2014} argues that the error of DFT can be further improved by using functionals considering full-range interactions and/or hybrid functionals with higher-order gradient terms. The CC is currently regarded as the gold standard for many quantum problems\cite{Bertels_2021}. The calculation of CC is based on a solution of the ground state of the system and is further stacked with a series of cluster operators to approximate the interactions between all electrons. These cluster operators can be extended from single and/or double to infinite order to increase the accuracy\cite{Zhang_2019}. 
Although DFT and CC have completely different theoretical frameworks, they share the same idea: adding a term to describe the full-range and complex many-body interactions can further improve the accuracy of the solutions.

\begin{figure}
    \centering
    \includegraphics[width=1.0\linewidth]{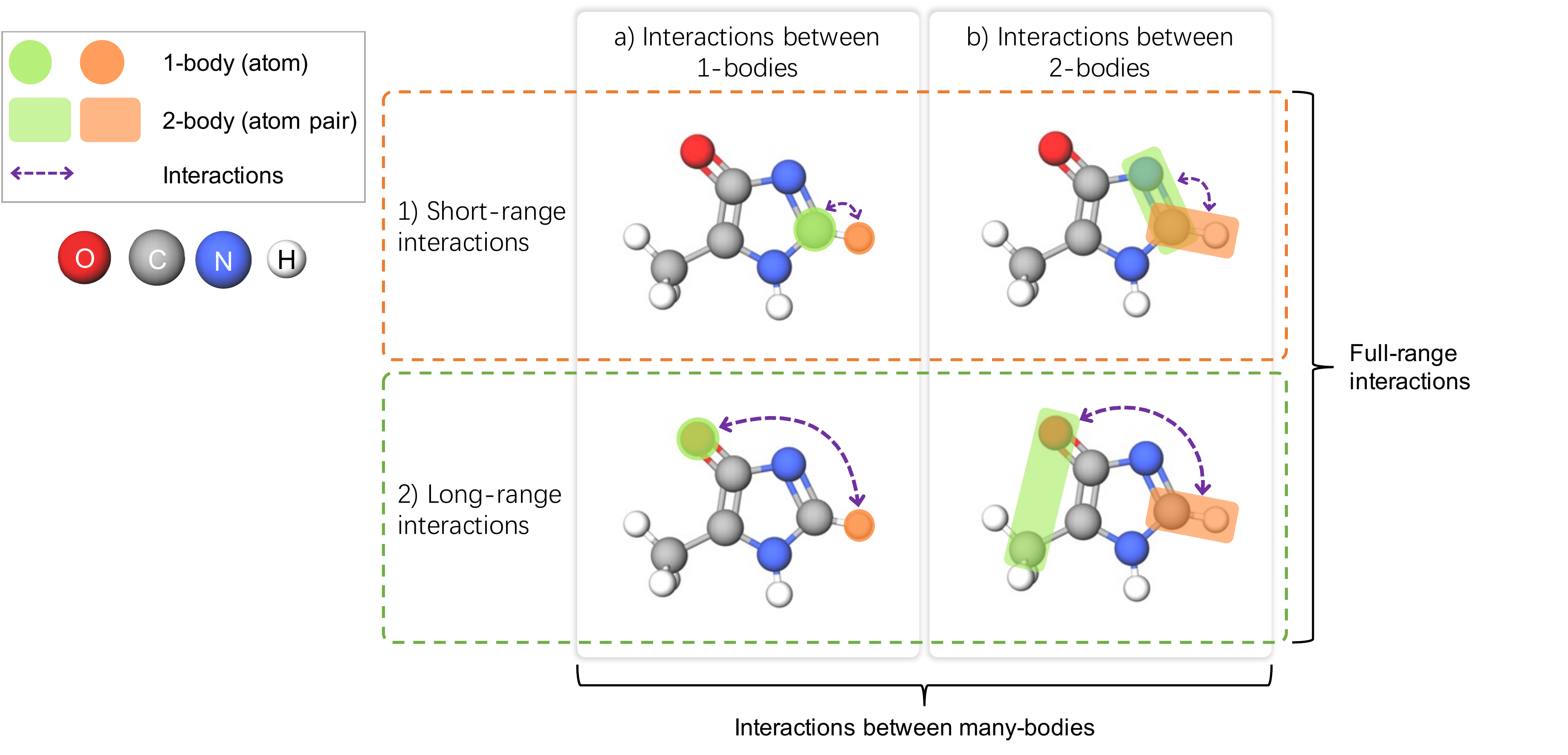} 
    \caption{Demonstration of many-body interactions and full-range interactions for molecular modeling. Many-body interactions include not only (a) the interactions between the 1-bodies (atoms) but also (b) the interactions between the higher-order many-bodies, e.g., 2-bodies (atom pairs). Full-range interactions are the interactions between any two many-bodies, including both (1) short-range interactions, e.g., interactions between atoms that are spatially close or connected by chemical bonds, and (2) long-range interactions, i.e., many-bodies that are not directly connected.}
    \label{fig:network_summary}
\end{figure}

Owing to the great success of deep learning methods in broad fields, many studies attempted to migrate widely used deep learning models to the field of molecular property prediction. Mainstream methods treat the compound as a graph and use a graph neural network (GNN) to model the interactions between nodes (nodes generally refer to atoms rather than electrons). Most of the methods\cite{DBLP:conf/iconip/DanelSTSSSM20:SGCN,DBLP:journals/corr/abs-2108-03348:EGT,DBLP:journals/corr/abs-2006-07739:DeeperGCN,DBLP:journals/corr/abs-2107-08773:CoMPT,DBLP:conf/nips/SchuttKFCTM17:SchNet} only take consideration of the interactions between atoms ($1$-bodies) in the molecule, as shown in column (a) in Figure~\ref{fig:network_summary}. Meanwhile, as demonstrated in column (b) in Figure~\ref{fig:network_summary}, several advanced works\cite{DBLP:conf/iclr/KlicperaGG20:DimeNet,DBLP:conf/icdm/ShuiK20:HMGNN,schutt2021equivariant:painn} start to incorporate interactions between atom pairs to utilize angular information for the purpose of geometric learning, which can be seen as interactions between $2$-bodies. Especially, our previous work GEM\cite{fang2022geometry:GEM} has demonstrated that interactions between the chemical bonds can also benefit the general molecular property prediction tasks. However, the studies that consider many-bodies mainly focus on the short-range interactions between the many-bodies, e.g., the interactions between the atoms that are close in space or connected by chemical bonds, as shown in row (1) in Figure~\ref{fig:network_summary}. That means the long-range interactions, e.g., the many-bodies that are not directly connected (as shown in row (2) in Figure~\ref{fig:network_summary}), are disregarded. Although increasing deep learning methods\cite{DBLP:journals/corr/abs-2002-08264:MAT,DBLP:journals/corr/abs-2012-09699:GT,DBLP:journals/corr/abs-2107-08773:CoMPT,DBLP:journals/corr/abs-2108-03348:EGT,DBLP:journals/corr/abs-2207-02505:TokenGT} have verified that directly modeling the full-range interactions (including the short-range and long-range interactions) between the atoms (1-bodies) can effectively enhance the accuracy for property prediction, there is still no investigation in how to design a model architecture that can comprehensively integrate the full-range many-body interactions. We argue that the consideration of the full-range many-body interactions is indispensable because this is not only an inherent requirement when solving the Schrödinger equation but has also been adopted by many computational chemistry methods as a basis to improve accuracy. Additionally, thoroughly describing the full-range many-body interactions is computationally expensive, especially when there are many atoms in the molecules, which is a huge challenge for the design of the learning mechanism.

To this end, we propose a novel molecular modeling framework, namely GEM-2, to comprehensively model the full-range many-body interactions for property prediction. GEM-2 introduces multiple tracks, where the $m^{th}$-track learns the interactions between the $m$-bodies. For each track, a novel and efficient many-body axial attention mechanism is designed to capture the full-range interactions, i.e., the interactions between any two many-bodies. The designed mechanism approximates the effect of the way that directly models full-range interactions through stacking axial attentions on multiple axes. We also exchange messages across different tracks, and in this way, the many-bodies with different orders can exploit the knowledge of each other.

To verify the effectiveness of GEM-2, we compare it with several competitive baselines on the tasks of quantum chemistry and drug discovery. In general, GEM-2 significantly outperforms the previous SOTA (state-of-the-art) methods on these tasks. Extensive ablation studies also demonstrate the advantages of incorporating the full-range many-body interactions for molecular property prediction. 

Our contributions can be summarized as follows:
\begin{itemize}
    \item We investigate the importance of full-range many-body interactions for molecular modeling and incorporate them in property prediction.
    \item A novel network architecture with multiple tracks is proposed to describe the full-range many-body interactions, and especially an efficient many-body axial attention mechanism is designed to model the full-range interactions.
    \item GEM-2 significantly outperforms the competitive baselines on quantum chemistry and drug discovery benchmarks. Extended ablation studies verify the contributions of full-range many-body interactions for molecular modeling.
\end{itemize}

\section{Methodology}

\begin{figure}[t]
    \begin{subfigure}{1.0\columnwidth}
      \centering
        \includegraphics[width=1.0\linewidth]{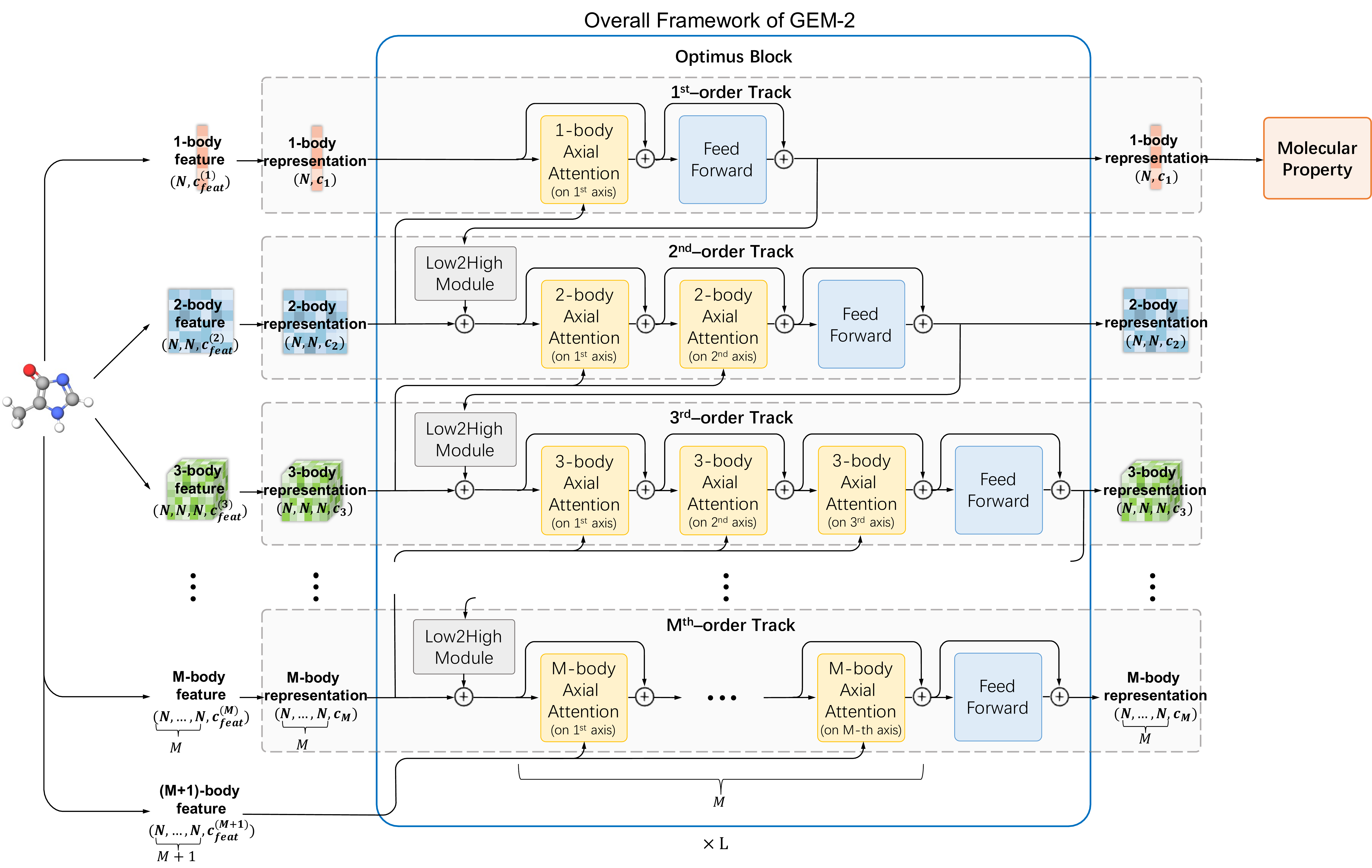}  
        \caption{The overall framework of GEM-2. First, a molecule is described by the representations of many-bodies of multiple orders. Then, Optimus blocks are designed to update the representations. Each Optimus block contains $M$ tracks, and the $m$-th track contains a stack of many-body axial attentions to model the full-range interactions between the $m$-bodies. The many-body axial attentions and the Low2High module also play the roles of exchanging messages across the tracks. Finally, the molecular property prediction is made by pooling over the $1$-body representations.}
        \label{fig:optimus_framework}
    \end{subfigure}
    
    \begin{subfigure}{1.0\columnwidth}
      \centering
        \includegraphics[width=1.0\linewidth]{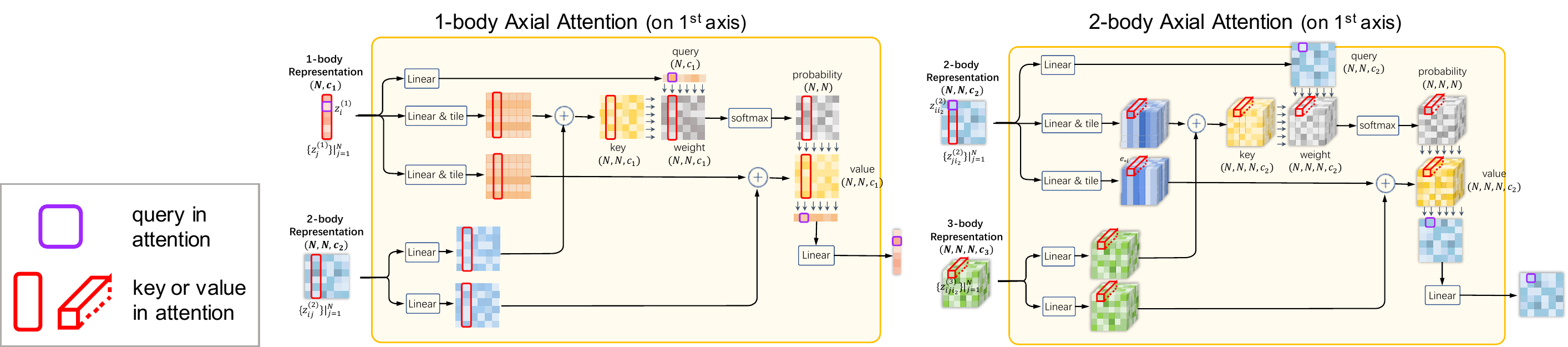}  
        \caption{Examples of the detailed model architectures of many-body axial attentions. Many-Body Axial Attentions regards the representations of the many-bodies as the queries, keys, and values of the attention mechanism. The knowledge of the many-bodies of higher-order are is also introduced for further improvement.}
        \label{fig:attention_framework}
    \end{subfigure}
    \caption{Demonstration of GEM-2.}
\end{figure}

\subsection{Overall Framework of GEM-2}
To start with, we define a $m$-body in a molecule as a group of $m$ atoms that acts as a whole, denoted as $(i_1,\cdots,i_m)$, with $i_1,...,i_m$ representing the indexes of the atoms in the molecule. $m$ is also called the order of that many-body. Aiming to incorporate the full-range many-body interactions into molecular modeling, we define a molecule with $N$ atoms as $\{\mathcal{V}^{(m)}\}|_{m=1}^{M+1}$. $\mathcal{V}^{(m)} = \{(i_1,\cdots,i_m)\}|_{1 \leq i_1,\cdots,i_m \leq N}$ is a set that contains all possible $m$-bodies of the molecule, i.e., the full permutation of $m$ atoms, and the size of $\mathcal{V}^{(m)}$ is $N^m$. Especially, $\mathcal{V}^{(1)}$ represents the atom set, $\mathcal{V}^{(2)}$ represents the set of atom pairs, and $\mathcal{V}^{(3)}$ represents the set of atom triplets. We attempt to learn the interactions between the $1^{st}$-bodies in $\mathcal{V}^{(1)}$, between the $2^{nd}$-bodies in $\mathcal{V}^{(2)}$, $\cdots$, and between the $M^{th}$-bodies in $\mathcal{V}^{(M)}$, respectively. Note that the information of the $(M+1)^{th}$-bodies $\mathcal{V}^{(M+1)}$ are additionally exploited to assist the modeling of many-body interactions. 

The input features of a $m$-body $(i_1,\cdots,i_m)$ and a $m$-body set $\mathcal{V}^{(m)}$ are denoted as $\bold{x}^{(m)}_{i_1,\cdots,i_m} \in \mathbb{R}^{c_{\text{feat}}^{(m)}}$ and $\bold{X}^{(m)} \in \mathbb{R}^{N^{m} \times c_{\text{feat}}^{(m)}}$ respectively, where $c_{\text{feat}}^{(m)}$ is the feature size of the $m$-body. 
For example, $\bold{X}^{(1)}$ contains the features of single atoms, such as atom type; $\bold{X}^{(2)}$ contains the features of atom pairs, such as chemical bond and spatial distance between two atoms; and $\bold{X}^{(3)}$ contains the features of atom triplets, such as angle formed by three atoms. (Please refer to Table~\ref{tab:features} in the Appendix for a detailed feature list.)

Modeling the full-range many-body interactions has been proven effective by the classical computational chemical methods for describing molecules. We define the full-range many-body interactions of the $m^{th}$-order as the interactions between any two $m$-bodies $(i_1,\cdots,i_m)$ and $(i'_1,\cdots,i'_m)$ in $\mathcal{V}^{(m)}$. Since $\mathcal{V}^{(m)}$ contains $N^{m}$ $m$-bodies, the full-range many-body interactions, i.e., $\{((i_1,\cdots,i_m), (i'_1,\cdots,i'_m))\}|_{1 \leq i_1,\cdots,i_m,i'_1,\cdots,i'_m \leq N}$, contains $N^{2m}$ interactions in theory. Consequently, directly learning the full-range many-body interactions is computationally expensive, especially when $N$ and $m$ are growing.

% In order to efficiently consider the full-range many-body interactions, we proposed a novel molecular modeling framework, called GEM-2, to comprehensively integrates the many-body set. The overall architecture of GEM-2 is demonstrated in Figure~\ref{fig:optimus_framework}. GEM-2 contains $L$ stacked Optimus blocks for molecular modeling, where Optimus is designed to learn the full-range interactions between the many-bodies with the same order as well as the interactions between the many-bodies with different orders. More concretely, we use $\bold{Z}^{(m)} \in \mathbb{R}^{N^m \times c_m}$ and $\bold{z}^{(m)}_{i_1,\cdots,i_m} \in \mathbb{R}^{N^m \times c_m}$ to denote representations of the $m$-body set and the $m$-body $(i_1,\cdots,i_m)$, respectively, where $c_m$ is the hidden size, and Optimus is used to update the many-body representations \liu{$\set{\bold{Z}^{(m)}}|_{m=1}^{M}$ up to the order $M$}. The transformation of the input features $\{\bold{X}^{(m)}\}|_{m=1}^{M}$ are taken as the initial representations. After that, the representations produced by the last Optimus block are utilized for molecular property prediction. The architecture details will be introduced in the following sections and the overall algorithm \ref{alg:gem-2} of GEM-2 can be found in the Appendix.

To effectively and also efficiently model the full-range many-body interactions, we propose a novel molecular modeling framework, namely GEM-2. The overall architecture of GEM-2 is demonstrated in Figure~\ref{fig:optimus_framework}. GEM-2 contains $L$ stacked Optimus blocks to iteratively update the many-body representations $\set{\bold{Z}^{(m)}}|_{m=1}^{M}$. $\bold{Z}^{(m)} \in \mathbb{R}^{N^m \times c_m}$ and $\bold{z}^{(m)}_{i_1,\cdots,i_m} \in \mathbb{R}^{c_m}$ denote representations of the $m$-body set $\mathcal{V}^{(m)}$ and a $m$-body $(i_1,\cdots,i_m)$, respectively, where $c_m$ is the hidden size of the representations. The Optimus blocks are designed to learn the full-range interactions between the many-bodies with the same order and also transfer the messages between the many-bodies with different orders. First, the transformation of the input features $\{\bold{X}^{(m)}\}|_{m=1}^{M}$ are taken as the initial representations $\{\bold{Z}^{(m)}\}|_{m=1}^{M}$. The Optimus blocks update the representations, and finally, the representations produced by the last Optimus block are utilized for molecular property prediction. The architecture details will be introduced in the following subsections.

\subsection{Optimus Block}

\begin{figure}
\centering
  \includegraphics[width=1.0\linewidth]{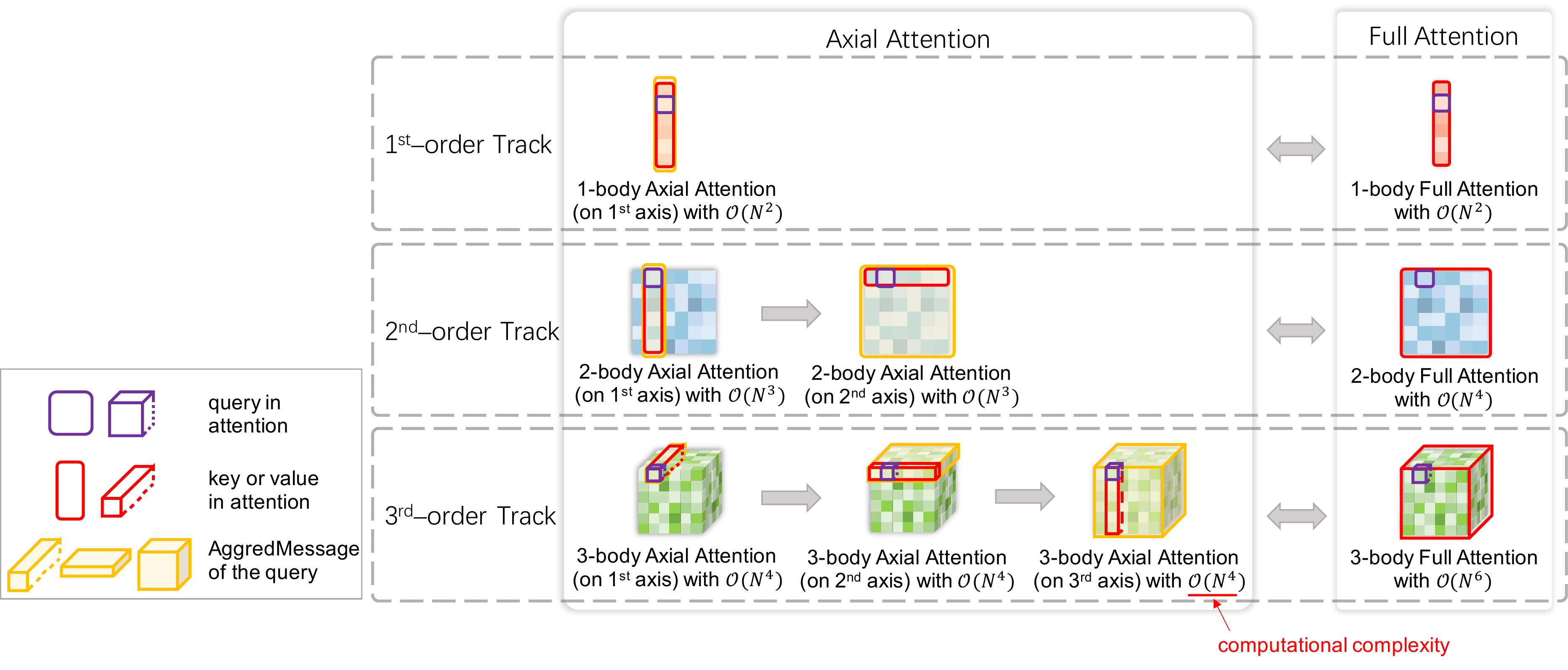}  
  \caption{Illustration of the principle of Many-body Axial Attention for full-range interaction modeling. The many-body representations of order $m$ in GEM-2 are organized as a multi-dimensional tensor $\bold{Z}^{(m)} \in \mathbb{R}^{N^m\times c_m}$, with each element $\bold{z}^{(m)}_{i_1,\cdots,i_m}$ in the tensor representing a $m$-body $(i_1,\cdots,i_m)$. For full-range interactions shown on the right, each many-body (element) of order $m$ collects the messages from all the many-bodies of the same order (all elements in the tensor). While for axial attention shown on the left, the target many-body (marked as purple) gradually aggregates the messages (marked as yellow) from all the many-bodies through $m$ stacked axial attention on multiple axes, approximating the effect of full attention that collects the messages of all the many-bodies in one shot with a much lower computational cost.}
  \label{fig:axial_demo}
\end{figure}

As shown in Figure~\ref{fig:optimus_framework}, an Optimus block consists of $M$ tracks of different orders where the $m^{th}$-order track updates $m$-body representations $\bold{Z}^{(m)}$. The $m^{th}$-order track captures the interactions between the $m$-, $(m-1)$- and $(m+1)$-bodies, i.e. $\bold{Z}^{(m)}$, $\bold{Z}^{(m-1)}$, and $\bold{Z}^{(m+1)}$. 
%We design a novel Many-body Axial Attention module to efficiently learn the full-range interactions within the $m$-bodies. Many-body Axial Attention focuses on modeling the full interactions of the $m$-body pairs that shares $m-1$ atoms, rather than indiscriminately modeling the interactions of all $N^{2m}$ $m$-body pairs at the same time. In this way, we can approximately model the full-range many-body interactions with a relatively low computational cost by stacking the axial attentions. 
A novel Many-body Axial Attention module is designed to efficiently learn the full-range interactions within the $m$-bodies. The stacked Many-body Axial Attention on multiple axes can approximately model the full-range interactions at a much lower computational cost than directly modeling the full-range interactions of all the $m$-body pairs at one shot.
%Secondly, by leveraging Many-body Axial Attention module and a Low2High module, the $m^{th}$-order track is also able to model the interactions of many-bodies with different orders by fusing messages from the high-order and lower-order track, i.e., the $(m+1)$- and $(m-1)$-th-order track.
Then, the Many-body Axial Attention module and the Low2High module aggregate representations from the high-order and lower-order track, i.e., the $(m+1)^{th}$- and $(m-1)^{th}$-order tracks, to exchange the messages across the tracks of different orders.
%\liu{Note that for the $1^{st}$-order track, the input $\bold{Z}^{(0)}$ is not needed, and for the $m^{th}$-order track, we set the input $\bold{Z}^{(M+1)}$ with $\bold{X}^{(M+1)}$. }

% As shown in Figure~\ref{fig:optimus_framework}, an Optimus block consists of $M$ tracks where the $m$-th track in Optimus mainly learns the $m$-body interactions by considering the $m$-th bodies, $(m-1)$-th bodies, and $(m+1)$-th bodies. The $m$-th track leverages Many-body Axial Attentions to efficiently learns the full-range interactions between the $m$-bodies. The Many-body Axial Attentions focus on modeling the interactions of those $m$-body pairs sharing $m-1$ atoms rather than indiscriminately model the interactions of all the $N^{2m}$ $m$-body pairs, i.e., $(\mathcal{V}^{(m)}, \mathcal{V}^{(m)})$. In this way, we can approximately model the full-range many-body interactions with a low computational cost. 
% The $m$-th track also takes advantage of the knowledge from the high-order track and the lower-order track, i.e., the $(m+1)$-th track and the $m-1$-th track, to deduce the interactions of the many-bodies with different orders through the axial attentions and a Low2High (L2H) module, respectively.

\subsubsection{Many-body Axial Attention}
Many-body Axial Attention is designed to efficiently learn the full-range many-body interactions of the same order as well as incorporating the message from the higher-order track. Note that the many-body representations of order $m$ in GEM-2 are organized as a multi-dimensional tensor $\bold{Z}^{(m)} \in \mathbb{R}^{N^m\times c_m}$, with each element $\bold{z}^{(m)}_{i_1,\cdots,i_m}$ in the tensor representing a $m$-body $(i_1,\cdots,i_m)$. Since the full-range many-body interactions, i.e., $\{((i_1,\cdots,i_m), (i'_1,\cdots,i'_m))\}|_{1 \leq i_1,\cdots,i_m,i'_1,\cdots,i'_m \leq N}$, contains $N^{2m}$ interactions, directly modeling these interactions in one shot is computational expensive. Inspired by axial attention\cite{ho2019axial}, we apply attention along a single axis of the tensor and run multiple times on different axes, learning the interactions of $m$-body pairs that share $m-1$ atoms, rather than applying attention to the flattened tensor elements. 

The illustration of the principle of Many-body Axial Attention for full-range interaction modeling is shown in Figure~\ref{fig:axial_demo}. For the sake of simplicity, we take the $m^{th}$-order track and a target $m$-body $(i_1, \cdots, i_m)$ for example. In order to learn the interactions between the target $m$-body to all the $m$-bodies in $\mathcal{V}^{(m)}$ (all the elements in the tensor), the original full attention (as shown on the right of Figure~\ref{fig:axial_demo}) takes that target $m$-body as the query of the attention mechanism, and all the $m$-bodies in $\mathcal{V}^{(m)}$ as keys and values of the attention mechanism. While for our proposed Many-body Axial Attention (as shown on the left of Figure \ref{fig:axial_demo}), we leverage $m$ stacked axial attentions that operate on multiple axes (from the $1^{st}$ axis to the $m^{th}$ axis) to approximate the effect of full attention. 
The axial attention on the $k^{th}$ axis takes the target $m$-body as the query of the attention mechanism (marked by purple in Figure \ref{fig:axial_demo}), and takes the $m$-bodies $\{(i_1, \cdots, i_{k-1}, j, i_{k+1}, \cdots, i_m)\}|_{j=1}^{N}$ (the elements marked by red in Figure \ref{fig:axial_demo}) as keys and values of the attention mechanism, which shares $(m-1)$ atoms (i.e. atom $i_1, \cdots, i_{k-1}, i_{k+1}, \cdots, i_m$) with the target $m$-body except atom $i_k$. 

%The axial attention on the $k^{th}$ axis calculates the attention on the $k^{th}$ axis of the representation tensor. It takes the target $m$-body as the query of the attention mechanism, and takes the $m$-bodies $\{(i_1, \cdots, i_{k-1}, j, i_{k+1}, \cdots, i_m)\}|_{j=1}^{N}$ as keys and values of the attention mechanism, which shares $(m-1)$ atoms (i.e. atom $i_1, \cdots, i_{k-1}, i_{k+1}, \cdots, i_m$) with the target $m$-body except atom $i_k$. 

By stacking the axial attentions from the $1^{st}$ axis to the $m^{th}$ axis, the target $m$-body can gradually aggregates the messages from all the $m$-bodies in $\mathcal{V}^{(m)}$. More specifically, in the $m$-body Axial Attention on $k^{th}$ axis, the target $m$-body is capable of aggregating the messages from many-bodies in the set which we define as $\text{AggredMessage}^{(k)}(i_1, \cdots, i_m)=\{(i'_1, \cdots, i'_k, i_{k+1}, \cdots, i_m)\}|_{1 \leq i'_1, \cdots, i'_k \leq N}$ (marked by yellow in the figure). The deduction is as follows:
\begin{linenomath}\begin{align}
    \text{AggredMessage}^{(0)}(i_1, \cdots, i_m) =& \{(i_1, \cdots, i_m)\}, \nonumber \\
    \text{AggredMessage}^{(k)}(i_1, \cdots, i_m) =& \cup_{i'_{k}=1}^{N}\text{AggredMessage}^{(k-1)}(i_1, \cdots, i_{k-1}, i'_{k}, i_{k+1}, \cdots, i_m) \nonumber \\
    =& \cup_{i'_{k}=1}^{N}\cup_{i'_{k-1}=1}^{N}\text{AggredMessage}^{(k-2)}(i_1, \cdots, i_{k-2}, i'_{k-1}, i'_{k}, i_{k+1}, \cdots, i_m) \nonumber \\
    & \cdots \nonumber \\
    =&\{(i'_1, \cdots, i'_k, i_{k+1}, \cdots, i_m)\}|_{1 \leq i'_1, \cdots, i'_k \leq N}.
\end{align}\end{linenomath}

%More specifically, at the axial attention on the $1^{st}$ axis, the target $m$-body collects all the messages from the $m$-bodies $\{(i_1, \cdots, i_m)\}|_{1 \leq i_1 \leq N}$. At the axial attention on the $2^{nd}$ axis, the target $m$-body collects the messages from $\{(i_1, \cdots, i_m)\}|_{1 \leq i_1, i_2 \le N}$, taking the $m$-bodies $\{(i_1, \cdots, i_m)\}|_{1 \leq i_2 \leq N}$ as intermediary that have already collected the corresponding messages at the axial attention on the $1^{st}$ axis. Similarly, at the axial attention on the $k^{th}$ axis, the target $m$-body can aggregate all the messages from $\{(i_1, \cdots, i_m)\}|_{1 \leq i_1, \cdots, i_k \leq N}$ (marked by yellow color in the figure)\todo{modify the figure color} by taking the $m$-bodies $\{(i_1, \cdots, i_m)\}|_{1 \leq i_k \leq N}$ as intermediary. 

Consequently, in the $m$-body Axial Attention on the $m^{th}$ axis, the target $m$-body can aggregate the messages from all the $m$-bodies in $\mathcal{V}^{(m)}=\{(i'_1,\cdots,i'_m)\}|_{1 \leq i'_1,\cdots,i'_m \leq N}=\text{AggredMessage}^{(m)}(i_1, \cdots, i_m)$. Computationally, it requires $\mathcal{O}(N^{m+1})$ to calculate one axial attention on a single axis and $\mathcal{O}(mN^{m+1})$ for the total $m$ stacked axial attentions. This is much more efficient than the original full attention which requires $\mathcal{O}(N^{2m})$. We provide the further discussion about the memory and computation cost in Section \ref{sec:memory_computation_cost} of Appendix.
% First, modeling the interactions of all the $N^{2m}$ $m$-body pairs, i.e., $(\mathcal{V}^{(m)}, \mathcal{V}^{(m)})$, is computational expensive, especially when the number of atoms $N$ is large and the order $m$ is high. We aim to design a technique that can approximate the full-range many-body interactions of order $m$ with a relatively low computational cost. We concentrate on the interactions of the $m$-body pairs that share $m-1$ atoms, and axial attention mechanism\cite{} is introduced to learn such interactions. For the $m$-th track, $m$ axial attentions on the $1$-st to the $m$-th, respectively, are stacked to approximate the full-range interactions. The mechanism of the stacked axial attention for many-body interaction modeling is demonstrated in Figure~\ref{}. Each element in the vectors/matrices/tensors denotes a many-body. 
To formalize Many-body Axial Attention, we take $m$-body axial attention on the $k^{th}$ axis as an example:
\begin{linenomath}\begin{align}
    \bold{q}_i&=\text{Linear}(\bold{z}^{(m)}_{i_1,\cdots,i_{k-1},i,i_{k+1},\cdots,i_m}), \nonumber \\ 
    \bold{k}_j&=\text{Linear}(\bold{z}^{(m)}_{i_1,\cdots,i_{k-1},j,i_{k+1},\cdots,i_m}), \nonumber \\
    \bold{v}_j&=\text{Linear}(\bold{z}^{(m)}_{i_1,\cdots,i_{k-1},j,i_{k+1},\cdots,i_m}), \nonumber \\ 
    \alpha_{ij}&=\text{softmax}(\bold{q}_i^T \bold{k}_j), \nonumber \\ \bold{o}_i&=\sum_{j}\alpha_{ij}\bold{v}_j,
\end{align}\end{linenomath}
where $\bold{q}_i$, $\bold{k}_j$ and $\bold{v}_j$ are the query, key, and value of the attention mechanism, respectively. The $m$-body representation $\bold{z}^{(m)}_{i_1,\cdots,i_{k-1},i,i_{k+1},\cdots,i_m}$ is then updated by the output representation $\bold{o}_i$ by the residual connection. 

% A $m$-th body (an element in the tensor) can receive the messages from all the $m$-th bodies in $\mathcal{V}^{(m)}$ (all the elements in the tensor) through the $m$ stacked axial attention\todo{revise the figure to better describe}. The stacked axial attentions effectively reduce the computational cost of $m$-th body full-range interaction modeling from $\mathcal{O}(N^{2m})$ to $\mathcal{O}(mN^m)$.

In addition, to incorporate the knowledge from the track of higher order, we further reform the axial attention for better molecular modeling. For the two $m$-bodies $(i_1,\cdots,i_{k-1}, i, i_{k+1},\cdots, i_m)$ and $(i_1,\cdots,i_{k-1}, j, i_{k+1}, \cdots, i_m)$ that share $(m-1)$ atoms, their interaction involves $(m+1)$ atoms. Those $(m+1)$ atoms as a whole can be regarded as a $(m+1)$-body $(i_1,\cdots,i_{k-1}, i, j, i_{k+1}, \cdots, i_m)$. We fuse the representations $\bold{z}^{(m+1)}_{i_1,\cdots,i_{k-1},i,j,i_{k+1},\cdots,i_m}$ of the $(m+1)$-body into the axial attention to incorporate the knowledge from the higher-order track. The representation of the corresponding $(m+1)$-body is taken as the additional keys and values of the attention. The $m$-body Axial Attention on the $k^{th}$ axis is redefined as
\begin{linenomath}\begin{align}
    \bold{q}_i&=\text{Linear}(\bold{z}^{(m)}_{i_1,\cdots,i_{k-1},i,i_{k+1},\cdots,i_m}), \nonumber \\ 
    \bold{k}_j&=\text{Linear}(\bold{z}^{(m)}_{i_1,\cdots,i_{k-1},j,i_{k+1},\cdots,i_m}), \quad \bold{k}_{ij}=\text{Linear}(\bold{z}^{(m+1)}_{i_1,\cdots,i_{k-1},i,j,i_{k+1},\cdots,i_m}), \nonumber \\
    \bold{v}_j&=\text{Linear}(\bold{z}^{(m)}_{i_1,\cdots,i_{k-1},j,i_{k+1},\cdots,i_m}), \quad \bold{v}_{ij}=\text{Linear}(\bold{z}^{(m+1)}_{i_1,\cdots,i_{k-1},i,j,i_{k+1},\cdots,i_m}),\nonumber \\ 
    \alpha_{ij}&=\text{softmax}(\bold{q}_i^T \bold{k}_{j} + \bold{q}_i^T \bold{k}_{ij})=\text{softmax}(\bold{q}_i^T (\bold{k}_{j}+\bold{k}_{ij})), \nonumber \\ \bold{o}_i&=\sum_{j}\alpha_{ij}(\bold{v}_{j}+\bold{v}_{ij}),
\end{align}\end{linenomath}
where $\bold{k}_{ij}$ and $\bold{v}_{ij}$ are the additional keys and values corresponding to the $(m+1)$-body. That means both the $m$-th body $(i_1,\cdots,i_{k-1}, j, i_{k+1},\cdots, i_m)$ and the $(m+1)$-th body $(i_1,\cdots,i_{k-1}, i, j, i_{k+1}, \cdots, i_m)$ are served as the keys and values of the axial attention. Such a combination can further improve the capacity of axial attention in modeling many-body interactions. Examples of the detailed architectures of the many-body axial attentions are exhibited in Figure~\ref{fig:attention_framework}.

\subsubsection{Exchanging Messages across the Tracks of Different Orders}
Many-bodies of different orders describe a molecule from different levels. For example, the $1^{st}$-order focuses on the atom-level knowledge, while the higher-orders focus on the motif-level or even higher-level knowledge. Exchanging the knowledge between the tracks of different orders facilitates each track to understand and utilize the knowledge at multiple levels. 

Many-body Axial Attention has already built a bridge to transfer the messages from the higher-order track to the lower-order track. We further introduce a Low2High module to transfer the messages from the lower-order track to the higher-order track. We introduce the Low2High module to the $2^{nd}$- to $M^{th}$-order tracks. For the $2^{nd}$-order track, the element-wise outer product operation is used as the Low2High module:
\begin{equation}
    \bold{o}^{(2)}_{i_1,i_2} = \text{OuterProduct}(\text{Linear}(\bold{z}^{(1)}_{i_1}), \text{Linear}(\bold{z}^{(1)}_{i_2})),
\end{equation}
where $\bold{o}^{(2)}_{i_1,i_2}$ denotes the updated $2$-body representation.
While for the $m^{th}$-order track with $2 < m \leq M$, we utilize the element-wise addition operation as the Low2High module:
\begin{linenomath}\begin{align}
    \bold{o}^{(m)}_{i_1,\cdots,i_m} &= \sum_{k=1}^{m}\text{Linear}(\bold{z}^{(m-1)}_{i_1,\cdots, i_{k-1}, i_{k+1}, \cdots, i_m}),
    %\bold{o}^{(m)}_{i_1,\cdots,i_m} &= \underbrace{\text{Linear}(\bold{z}^{(m-1)}_{i_2,\cdots,i_m}) + \text{Linear}(\bold{z}^{(m-1)}_{i_1,i_3,\cdots,i_m}) + \cdots + \text{Linear}(\bold{z}^{(m-1)}_{i_1,\cdots,i_{m-1}})}_{m},
    %\bold{o}^{(m)}_{i_1,\cdots,i_m} &= \text{Linear}(\text{Activation}(\bold{o}^{(m)}_{i_1,\cdots,i_m})),
\end{align}\end{linenomath}
where $\bold{o}^{(m)}_{i_1,\cdots,i_m}$ denotes the updated $m$-body representation.

\subsection{Molecular Property Prediction}
The messages of the many-bodies can be exchanged across the tracks with different orders in the Optimus block. Through $L$ stacked Optimus blocks where $L$ is usually set to a value larger than $M$, the representations of the atoms ($1$-bodies) produced by the last Optimus block are able to integrate the information of the many-bodies with various orders and can be used for molecular property prediction. We first apply average pooling on the atom representations, $\bold{Z}^{(1)}$, to obtain the representation of the whole molecule. Then, the molecular representation is fed into a Multilayer Perceptron (MLP) to output the predicted properties.

\section{Results}
To comprehensively evaluate the performance of GEM-2 for molecular property prediction, we compare it with multiple baseline methods on two kinds of benchmarks: quantum chemistry and drug discovery. 
%For the quantum chemistry domain, PCQM4Mv2\cite{hu2021ogblsc} is one of the most largest molecular property prediction benchmarks so far, which is suitable for benchmarking the ability of different networks on molecular modeling. While for the drug discovery domain, we choose the virtual screening scenario, which not only require the modeling of molecules, but also need to inexplicitly infer the structures of the target protein. And LIT-PCBA\cite{tran2020lit-pcba} is a popular virtual screening dataset famous for its realism and unbiasedness.
Furthermore, we also conduct extensive ablation studies to investigate the impact of the full-range interactions and the orders of many-bodies. 

\subsection{Experimental Settings of GEM-2}
In all the experiments, GEM-2 contains 12 Optimus blocks. For computational efficiency, each Optimus block contains two tracks, i.e. the $1^{st}$-order track and the $2^{nd}$-order track, unless otherwise specified. That means we consider the $1$-bodies, $2$-bodies, and the $3$-bodies for molecular modeling. In order to gain the best performance, the hyper-parameters are slightly tuned, for example, we set $c_1=c_2=c_3=256$ for the quantum chemistry benchmark and $c_1=c_2=c_3=128$ for the drug discovery benchmark. While for each ablation study, in order to save computational consumption without losing fairness, we use smaller hidden sizes, and the hyper-parameters are fixed for different GEM-2 variants. For model optimization, we use Adam Optimizer\cite{DBLP:journals/corr/KingmaB14} to train GEM-2. Besides, exponential moving average (EMA)\cite{polyak1992acceleration:EMA} with a decay rate of 0.999 is exploited to smooth the model parameters for the sake of achieving more robust performance. The detailed settings are described in Section~\ref{sec:hyper_parameters} in Appendix.

The input features for GEM-2 can be generally divided into three types in our experiments: features for $1$-body describing atoms; features for $2$-body describing atom pairs, and features for $3$-body describing atom triplets. All these features are fetched from the cheminformatics tool RDKit (\url{https://www.rdkit.org}), including the Merck molecular force field (MMFF94)\cite{halgren1996merck} function to obtain simulated three-dimensional coordinates of atoms. We utilize the same feature set for all our experiments. Please refer to Appendix~\ref{sec:input_features} for the detailed feature list.

% For classification tasks, we use binary cross entory as the loss function and area under the receiver operating characteristic curve (ROC-AUC) as the evaluation metric, for which the higher the better. For multi-target task, different targets share the same encoder body but get their own heads.

\subsection{Quantum Chemistry Benchmark}
\label{sec:quantum}
\textbf{Dataset.}
PCQM4Mv2\cite{hu2021ogblsc} is a large-scale quantum chemistry dataset containing the DFT-calculated HOMO-LUMO energy gaps of 3,746,619 molecules, which is originally curated under the PubChemQC project\cite{nakata2017pubchemqc}. Open Graph Benchmark (OGB)\footnote{https://ogb.stanford.edu} split the dataset according to their PubChem ID (CID) into training, validation, test-dev, and test-challenge sets with a ratio of 90:2:4:4. The training set is used for training and the validation set is used for model selection. We report the performance of GEM-2 and other baseline methods on both validation and test-dev sets. Note that the results of test-dev set can only be obtained by submitting to the OGB server, thus some results on test-dev are not reported. Also, the test-challenge set has not been released by OGB so far.

\textbf{Baselines.}
We compare GEM-2 with multiple baseline methods. They can be roughly classified into three categories according to whether considering full-range interactions and many-body interactions: 1) \textit{w/o full-range \& w/o many-body}, 2) \textit{w/ full-range \& w/o many-body}, and 3) \textit{w/o full-range \& w/ many-body}. \textit{w/o full-range \& w/o many-body} methods are based on graph neural networks (GNNs) and only consider the atom-level interactions between the atoms connected by chemical bonds, including Graph Isomorphism Network(GIN)\cite{DBLP:conf/iclr/XuHLJ19}, GIN-virtual, Graph Convolutional Networks (GCN)\cite{DBLP:journals/corr/KipfW16}, and GCN-virtual, where GIN-virtual and GCN-virtual are the revised versions of GIN and GCN that utilize virtual node\cite{DBLP:conf/icml/GilmerSRVD17:virtual}. \textit{w/ full-range \& w/o many-body} methods are Transformer-style networks that directly model the interactions between all the atoms, including TokenGT\cite{DBLP:journals/corr/abs-2207-02505:TokenGT}, GRPE-Large\cite{park2022graph}, EGT\cite{DBLP:journals/corr/abs-2108-03348:EGT}, Graphormer\cite{ying2021transformers} and GPS\cite{DBLP:journals/corr/abs-2205-12454:GPS}. They only consider the atom-level interactions without considering the interactions between the many-bodies. \textit{w/o full-range \& w/ many-body} methods, including our previous work, GEM \cite{fang2022geometry:GEM}, take the many-body interactions into account but focus on the short-range interactions, e.g., interactions between the atoms connected by the chemical bonds.

%\todo{training time}
 
\textbf{Results.}
We regard the prediction of HOMO-LUMO energy gaps as a regression problem and apply L1 loss as the loss function and Mean Absolute Error (MAE) as the evaluation metric. 
The results of GEM-2 and the baseline methods are shown in Table~\ref{tab:PCQM4Mv2} (The lower, the better). The results of GCN, GIN, GCN-vritual, GIN-virtual, TokenGT, GRPE-Large, and EDT are collected from the PCQM4Mv2 leaderboard\footnote{https://ogb.stanford.edu/docs/lsc/leaderboards/\#pcqm4mv2} in OGB, the results of Graphormer and GPS are from their papers, and the results of GEM\cite{fang2022geometry:GEM} are obtained by adapting the open source code from the corresponding GitHub repositories to re-train the models on the PCQM4Mv2 dataset.
From the results, we can draw the following conclusions:
\begin{enumerate}
    \item Generally, GEM-2 significantly outperforms all the baseline methods that do not consider the full-range interactions or the many-body interactions for molecular modeling. GEM-2 achieves a relative improvement of $7.5\%$ and $6.5\%$ compared with the previous SOTA, i.e., EGT, on the validation and test-dev sets, respectively, demonstrating its superiority.
    \item The MAE scores of the methods that model the full-range interactions are lower than those of the methods that only focus on the short-range interactions (the interactions between the locally closed atoms or many-bodies), no matter whether considering the many-body interactions or not. The results indicate that directly modeling full-range interactions is more effective in molecular modeling, enhancing the property prediction accuracy.
    \item By comparing the methods that incorporate the many-body interactions for molecular modeling and the methods that only model the interactions between the atoms ($1$-bodies), we found that it is challenging for the methods that only model the interactions between the atoms to achieve further improvement, even though some of the baseline methods have already learned the full-range interactions through variant model architectures.
\end{enumerate}

\begin{table}
\caption{MAE scores (lower is better) of GEM-2 and multiple kinds of baselines on quantum chemistry dataset PCQM4Mv2.}
\centering
\label{tab:PCQM4Mv2}
\begin{threeparttable}
\begin{tabular}{l|l|ccc}
\toprule
    & Model              & Validation MAE $\downarrow$ & Test-dev MAE $\downarrow$ & \#Params \\
\midrule
\multirow{4}{*}{w/o full-range \& w/o many-body} 
    & GCN\cite{DBLP:journals/corr/KipfW16}                & 0.1379    & 0.1398  & 2.0M     \\
    & GIN\cite{DBLP:conf/iclr/XuHLJ19}                & 0.1185    & 0.1218  & 3.8M      \\
    & GCN-virtual\cite{DBLP:conf/icml/GilmerSRVD17:virtual}        & 0.1153    & 0.1152  & 4.9M     \\
    & GIN-virtual\cite{DBLP:conf/icml/GilmerSRVD17:virtual}        & 0.1083    & 0.1084   & 6.7M    \\
\midrule
\multirow{5}{*}{w/ full-range \& w/o many-body} 
    & TokenGT\cite{DBLP:journals/corr/abs-2207-02505:TokenGT}                 & 0.0910     & 0.0919  & 48.5M     \\
    & GRPE-Large\cite{park2022graph}                 & 0.0867     & 0.0876  & 118.3M     \\
    & Graphormer\cite{ying2021transformers}         & 0.0864    &    -\tnote{a}     & 48.3M     \\
    & GPS\cite{DBLP:journals/corr/abs-2205-12454:GPS}               & 0.0858    & -\tnote{a}    & 19.4M   \\
    & EGT\cite{hussain2021edge}               & 0.0857    & 0.0862 & 89.3M      \\
\midrule
\multirow{1}{*}{w/o full-range \& w/ many-body} 
    & GEM\cite{fang2022geometry:GEM}  & 0.0904    &    -\tnote{a}   & 32.4M       \\
\midrule
w/ full-range \& w/ many-body
    & GEM-2     & \textbf{0.0793}    & \textbf{0.0806} & 32.1M            \\
\bottomrule
\end{tabular} 
\begin{tablenotes}
    \tiny
    \item The SOTA results are shown in bold.
    \item[a] The result of the corresponding method on test-dev set is not reported by the OGB server.
\end{tablenotes}
\end{threeparttable}

% \begin{tablenotes}
%   \tiny
%   \item The SOTA results are shown in bold.
%   \item The cells in gray indicate the previous SOTA results.
% \end{tablenotes}
\end{table}

\subsection{Drug Discovery Benchmark}
\label{sec:drug}
\textbf{Dataset.}
We further verify the effectiveness of GEM-2 on the virtual screening task for drug discovery. LIT-PCBA\cite{tran2020lit-pcba} is a virtual screening dataset containing 15 protein targets, 9780 active compounds (positive samples), and 407,839 unique inactive compounds (negative samples) selected from high-confidence PubChem Bioassay data. Predicting the activities of the candidate compounds to a particular protein target can be regarded as a binary classification task. Due to the large prediction variance on the classification tasks with only a few positive samples, we only evaluate the methods on the targets (ALDH1, FEN1, GBA, KAT2A, MAPK1, PKM2, and VDR) with more than 150 active compounds. Following the previous work\cite{cai2022fp}, we split the samples of each protein target into the training and test sets at the ratio of 3:1 with asymmetric validation embedding (AVE) method\cite{wallach2018most}. The split dataset can be directly downloaded at the previous work's GitHub repository\footnote{https://github.com/idrugLab/FP-GNN}. Additionally, we reserve 1/9 samples in the training set as the validation set to select the best model and then evaluate the models' performance on the test set.

\textbf{Baselines.}
Similar to the experiments on PCQM4Mv2, we compare GEM-2 with multiple types of baselines:
(1) traditional machine learning methods, including Naive Bayes(NB)\cite{duda1973pattern}, support vector machine (SVM) \cite{cortes1995support}, random forest (RF) \cite{liaw2002classification} and extreme gradient boosting (XGBoost)\cite{chen2016xgboost}; (2) \textit{w/o full-range \& w/o many-body} methods, including GCN, Graph Attention Network (GAT) \cite{DBLP:journals/corr/abs-1710-10903}, and FP-GNN \cite{cai2022fp}; (3) \textit{w/ full-range \& w/o many-body} methods, including EGT and EGT\subtext{pretrain}\cite{DBLP:journals/corr/abs-2108-03348:EGT}; (4) \textit{w/o full-range \& w/ many-body} methods, including GEM and GEM\subtext{pretrain}\cite{fang2022geometry:GEM}. GEM-2 and GEM-2\subtext{pretrain} are our proposed methods. 
As the number of positive samples in LIT-PCBA is insufficient and may lead to over-fitting, we also implement the pretrained versions for methods EGT, GEM, and GEM-2 to alleviate the impact of over-fitting, marked with subscript ``pretrain''. For each method, we first pretrain it on the quantum chemistry benchmark PCQM4Mv2. Then, the pretrained model is finetuned on dataset LIT-PCBA. 

\textbf{Results.}
Since each target is taken as a binary classification task, binary cross entropy is used as the loss function to optimize the models. To compare the accuracy of GEM-2 and the baseline methods, we use ROC-AUC (area under the receiver operating characteristic curve) as the evaluation metric. The results of NB, SVM, RF, XGBoost, GCN, GAT, and FP-GNN are collected from \cite{cai2022fp}. The results of EGT, EGT\subtext{pretrain}, GEM, GEM\subtext{pretrain} are obtained by running the released code with hyper-parameter searching on dataset LIT-PCBA. We ran each experiment three times and reported the average AUC scores to reduce the experimental variance caused by the limited positive samples. From Table~\ref{tab:litpcba}, we can observe that GEM-2\subtext{pretrain} performs the best on 6 out of 7 targets, which is in line with the results on the quantum chemistry dataset, demonstrating its great potential in the drug discovery industry. Besides, the pretrained versions of EGT, GEM, and GEM-2 all gain further improvement compared with the corresponding no-pretraining versions. %Especially, the no-pretrain version of EGT suffers seriously from over-fitting.

\begin{table}
\caption{AUC scores (higher is better) of GEM-2 and the baseline methods on drug discovery dataset LIT-PCBA. }
\centering
\small
\label{tab:litpcba}
\begin{threeparttable}
\begin{tabular}{l|ccccccc|c}
\toprule
        & ALDH1 $\uparrow$      & FEN1 $\uparrow$       & GBA $\uparrow$        & KAT2A $\uparrow$      & MAPK1 $\uparrow$         & PKM2 $\uparrow$       & VDR $\uparrow$        & Average $\uparrow$    \\
\midrule
No. active & 7,168 & 369 & 166 & 194 & 308 & 546 & 884 & - \\
No. inactive & 137,965 & 355,402 & 296,052 & 348,548 & 62,629 & 245,523 & 355,388 & - \\

\midrule
NB\cite{duda1973pattern}\tnote{a}       & 0.693      & 0.876      & 0.709      & 0.659      & 0.686          & 0.684      & 0.804      & 0.730   \\

SVM\cite{cortes1995support}\tnote{a}      & 0.76       & 0.877      & 0.778      & 0.612      & 0.665          & 0.753      & 0.69       & 0.734    \\

RF\cite{liaw2002classification}\tnote{a}       & 0.741      & 0.657      & 0.599      & 0.537      & 0.579          & 0.581      & 0.644      & 0.620    \\

XGBoost\cite{chen2016xgboost}\tnote{a}  & 0.75       & 0.888      & 0.83       & 0.5        & 0.593          & 0.737      & 0.782      & 0.726  \\
\midrule

GCN\cite{DBLP:journals/corr/KipfW16:GCN}\tnote{a}      & 0.73       & 0.897      & 0.735      & 0.621      & 0.668            & 0.636      & 0.773      & 0.723   \\

GAT\cite{DBLP:journals/corr/abs-1710-10903}\tnote{a}     & 0.739      & 0.888      & 0.776      & 0.662      & 0.697          & 0.724      & 0.78       & 0.752   \\

FP-GNN\cite{cai2022fp}\tnote{a}   & 0.766      & 0.889      & 0.751      & 0.632      & \textbf{0.771}          & 0.732      & 0.774      & 0.759   \\

\midrule
EGT\cite{DBLP:journals/corr/abs-2108-03348:EGT}   &  0.725\subtext{(0.01)}      & 0.810\subtext{(0.05)}      & 0.529\subtext{(0.12)}      & 0.546\subtext{(0.01)}      & 0.675\subtext{(0.02)}      & 0.646\subtext{(0.04)}      & 0.740\subtext{(0.01)}      & 0.667    \\

EGT$_{\text{pretrain}}$\cite{DBLP:journals/corr/abs-2108-03348:EGT} &  0.787\subtext{(0.02)}      & 0.929\subtext{(0.01)}      & 0.754\subtext{(0.04)}      & 0.728\subtext{(0.01)}      & 0.753\subtext{(0.03)}      & 0.765\subtext{(0.02)}      & 0.807\subtext{(0.02)}      & 0.789   \\

\midrule

GEM\cite{fang2022geometry:GEM}   &  0.776\subtext{(0.003)}      & 0.933\subtext{(0.01)}      & 0.829\subtext{(0.01)}      & 0.632\subtext{(0.09)}      & 0.685\subtext{(0.02)}      & 0.735\subtext{(0.04)}      & 0.771\subtext{(0.02)}      & 0.766   \\

GEM\subtext{pretrain}\cite{fang2022geometry:GEM}  &  0.772\subtext{(0.01)}      & 0.914\subtext{(0.02)}      & 0.821\subtext{(0.02)}      & 0.740\subtext{(0.01)}      & 0.710\subtext{(0.02)}      & 0.746\subtext{(0.02)}      & 0.785\subtext{(0.01)}      & 0.784   \\
\midrule
GEM-2  &  0.770\subtext{(0.02)}      & 0.929\subtext{(0.01)}      & 0.819\subtext{(0.02)}      & 0.670\subtext{(0.02)}      & 0.715\subtext{(0.02)}      & 0.724\subtext{(0.03)}      & 0.805\subtext{(0.02)}      & 0.776   \\

GEM-2\subtext{pretrain}  &  \textbf{0.802\subtext{(0.002)}}      & \textbf{0.945\subtext{(0.003)}}      & \textbf{0.856\subtext{(0.02)}}      & \textbf{0.763\subtext{(0.01)}}      & 0.733\subtext{(0.01)}      & \textbf{0.782\subtext{(0.004)}}      & \textbf{0.823\subtext{(0.005)}}      & \textbf{0.815}    \\
\bottomrule
\end{tabular}
\begin{tablenotes}
    \tiny
    \item The SOTA results are shown in bold. Standard deviations are shown in brackets.
    \item[a] These results are collected from \cite{cai2022fp}, where standard deviations are not reported.
\end{tablenotes}
\end{threeparttable}
\end{table}

\subsection{Ablation Study}
In this section, we conduct extensive ablation studies on dataset PCQM4Mv2 to investigate the contributions of many-body interactions and full-range interactions for molecular modeling. We further analyze the impact of the many-body interactions with different orders and the long-range interactions with different levels. We report the MAE scores of the ablation versions of GEM-2 on the PCQM4Mv2 valid set. Note that we use a smaller hidden size for comparison in this subsection to save computational consumption without losing the fairness of comparison. Please refer to Appendix~\ref{sec:hyper_parameters} for the detailed settings. 
%Besides, additional experiments to analyze the impact of the architecture design of the Optimus block can be found in Appendix~\ref{sec:further_ablation}.

\begin{figure}[t]
    \centering
    \begin{subfigure}[b]{1.0\textwidth}
     \centering
     \includegraphics[width=0.8\textwidth]{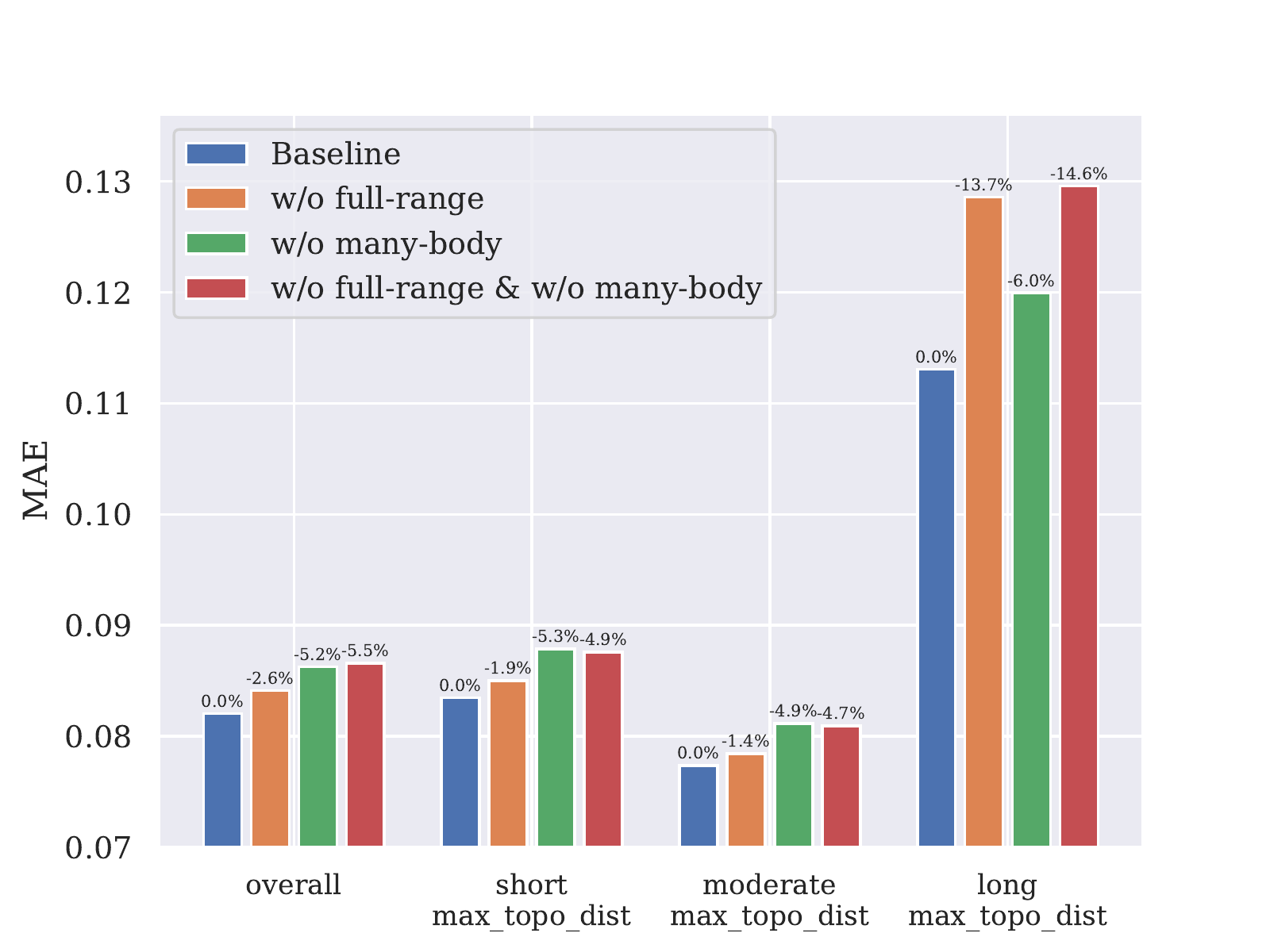}
     \caption{MAE scores of GEM-2 variants to analyze the contributions of many-body interactions and full-range interactions. Besides the \textit{overall} group, the test molecules are further divided into three groups with different $max\_topo\_dist$: (1) $short~max\_topo\_dist$, (2) $moderate~max\_topo\_dist$, and (3) $long~max\_topo\_dist$. The relative improvement compared to \textit{Baseline} is also shown at the top of each bar.}
     \label{fig:many_vs_long}
    \end{subfigure}
    \centering
     \begin{subfigure}[b]{0.48\textwidth}
         \centering
         \includegraphics[width=\textwidth]{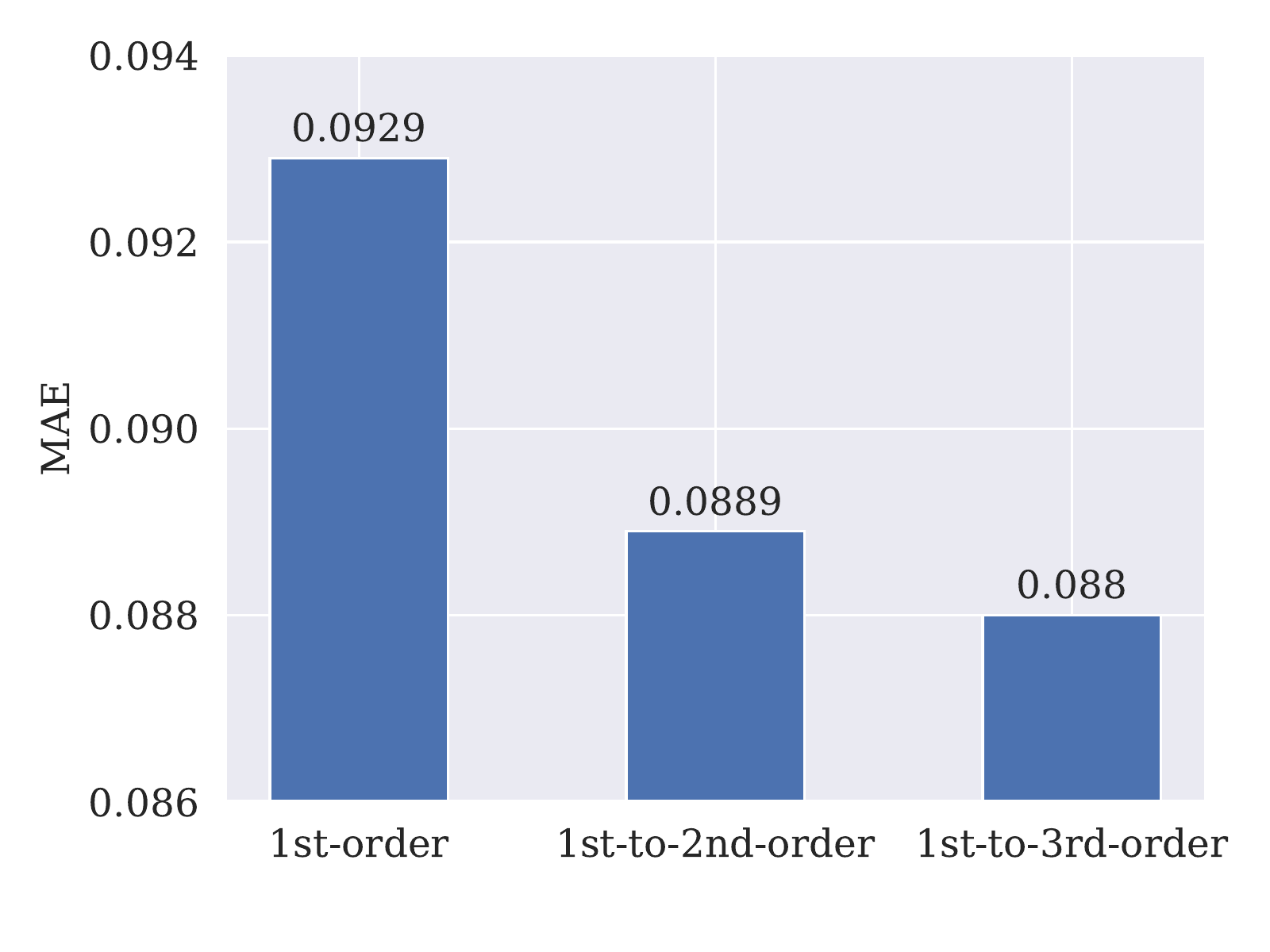}
         \caption{MAE scores of GEM-2 variants with different orders of many-bodies.}
         \label{fig:order}
     \end{subfigure}
     \hfill
     \begin{subfigure}[b]{0.48\textwidth}
         \centering
         \includegraphics[width=\textwidth]{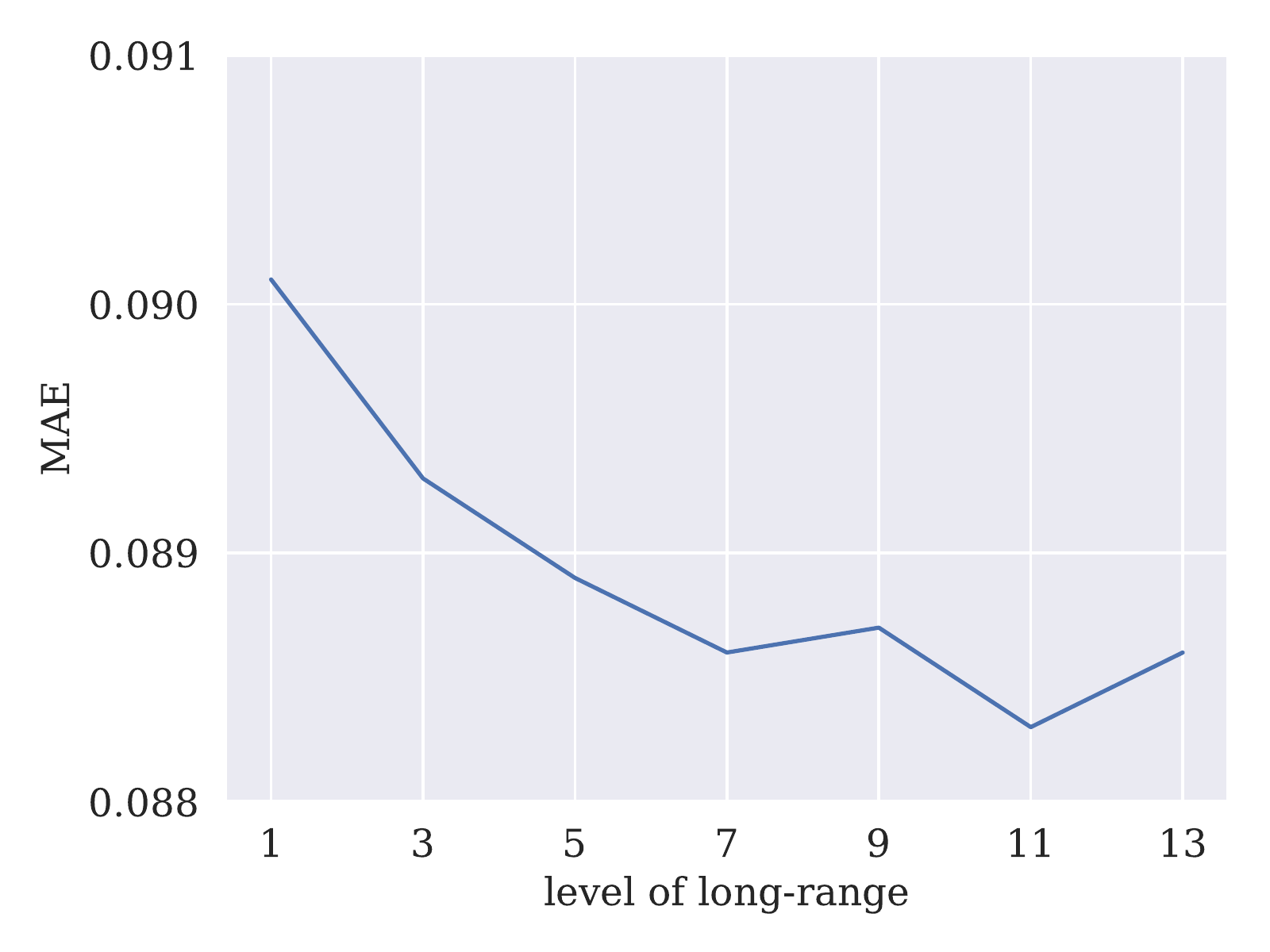}
         \caption{MAE scores of GEM-2 variants with different long-range levels.}
         \label{fig:hop}
     \end{subfigure}
\caption{Ablation studies of GEM-2 on PCQM4Mv2.}
\end{figure}

\subsubsection{Effects of Many-Body Interactions and Full-Range Interactions}
\label{sec:many_vs_long}
GEM-2 incorporates the many-body interactions and the full-range interactions as a whole for molecular modeling. We compare four variants of GEM-2: 1) \textit{Baseline} is the complete GEM-2 that models the full-range many-body interactions; 2) \textit{w/o full-range} variant only models the short-range interactions of those many-bodies, e.g., the interactions between those atoms that are connected by the chemical bonds; 3) \textit{w/o many-body} variant only uses the $1^{st}$ order track in Optimus block, i.e. only considers the interactions between the atoms ($1$-bodies); 4) \textit{w/o full-range \& w/o many-body} variant focuses on learning the short-range interactions between the atoms, which is a combination of \textit{w/o full-range} variant and \textit{w/o many-body} variant.

Besides, we suspect that the original GEM-2 has great advantages in modeling full-range interactions, especially long-range interactions. To verify this assumption, we further test molecules with different $max\_topo\_dist$ where $max\_topo\_dist$ stands for the maximal topological distance of any two atoms in the molecules. And the topological distance between two atoms refers to the minimal number of chemical bonds through which those atoms can be connected. Usually, the larger the maximum topological distance of a molecule, the greater the importance of long-range interaction modeling for that molecule. Besides the \textit{overall} group that contains all the test molecules, we further divide the molecules into three groups with different level of $max\_topo\_dist$: (1) $short~max\_topo\_dist$ group, with $max\_topo\_dist$ $\in [1,7]$, (2) $moderate~max\_topo\_dist$ group, with $max\_topo\_dist$ $\in [8,11]$, and (3) $long~max\_topo\_dist$ group, with $max\_topo\_dist$ $\in [12,\infty)$. 

%We further divide the molecules of PCQM4Mv2 valid set into multiple groups by different $max\_hop$, where $max\_hop$ of a molecule is the maximal topological hop distance (i.e. shortest path distance) between any two atoms. Four groups of molecules are evaluated in the experiments: 1) \textit{Overall} group contains all molecules from the valid set of PCQM4Mv2; 2) \textit{$max\_hop \in [0,8)$} group keeps molecules with small $max\_hop$ less than $8$; 2) \textit{$max\_hop \in [8,12)$} group keeps molecules with medium $max\_hop$ between $8$ (inclusive) and $12$ (exclusive); 3) \textit{$max\_hop \in [8,\text{inf})$} group keeps molecules with large $max\_hop$ larger than or equal to $12$.
We compare the MAE scores of multiple variants of GEM-2 in Figure~\ref{fig:many_vs_long}. In general, \textit{Baseline} outperforms the other variants on all the molecular groups. Especially, the complete GEM-2, i.e., \textit{Baseline}, shows excellent superiority on $long~max\_topo\_dist$ group. (1) Compared \textit{Baseline} with \textit{w/o full-range}, we can observe that \textit{Baseline} achieves a significant improvement of 13.7\% on $long~max\_topo\_dist$ group and a relatively slight improvement on $short~max\_topo\_dist$ group and $moderate~max\_topo\_dist$ group. Although methods only focusing on the local relations between the atoms can model those molecules with few long-range interactions, its modeling ability for large or complex molecules is unsatisfactory. (2) The improvement of many-body interaction modeling, by comparing \textit{Baseline} and \textit{w/o many-body}, is also significant and consistent (from 4.9\% to 6.0\%) on all the molecule groups. Many-body interaction modeling contributes a lot to molecular modeling, which is in line with the conclusion of the classical chemical computational methods and previously proposed deep-learning methods based on many-bodies.

% Both the full-range and many-body interactions can improve the capacity of GEM-2 for molecular modeling. The MAE of \textit{w/o full-range} variant is close to that of \textit{w/o full-range \& w/o many-body} and is lower than that of \textit{w/o many-body} variant. These results indicate that modeling many-body interactions are critical for molecular property prediction. The role of learning the full-range interactions is limited without many-body interaction modeling, while full-range interaction modeling based on many-body interaction modeling can achieve further improvement. The baseline version of GEM-2, \textit{baseline}, which captures the full-range many-body interactions for molecular modeling, works the best.

% \begin{table}[t]
% \caption{MAE of GEM-2 variants (many-body versus long-range interaction) on PCQM4Mv2. Lower is better.}
% \centering
% \small
% \label{tab:many_vs_long}
% \begin{tabular}{l|cccc}
% \toprule
%   & Short-range $2$-body interaction  & Full-range $2$-body interaction & Short-range many-body interaction & Baseline \\
% \midrule
% MAE & 0.0862 & 0.0858 & 0.0837 & 0.0819 \\
% \bottomrule
% \end{tabular}
% \end{table}

% \begin{figure}
%     \centering
%     \includegraphics[width=0.5\linewidth]{example-image-b}
%     \caption{Figure of visualization of isomerism samples}
%     \label{fig:isomerism}
% \end{figure}

\subsubsection{Effects of the Orders of Many-Bodies}
\label{sec:order}
We have demonstrated the overwhelming superiority over existing baseline methods in the previous subsections, where each Optimus block in GEM-2 involves two tracks for many-body modeling, i.e., $1^{st}$-to-$2^{nd}$-order tracks. We believe incorporating higher-order many-bodies can further enhance the capacity of molecular modeling. To investigate the effects of the orders of many-bodies for molecular modeling, we conduct ablation studies on dataset PCQM4Mv2 to compare the following variants of GEM-2: 1) \textit{$1^{st}$-order} containing only the $1^{st}$-order track; 2) \textit{$1^{st}$-to-$2^{nd}$-order} containing the $1^{st}$- and $2^{nd}$-order tracks; 3) \textit{$1^{st}$-to-$3^{rd}$-order} containing the $1^{st}$-, $2^{nd}$- and $3^{rd}$-order tracks.

The MAE scores of the GEM-2 variants with different orders of many-bodies are shown in Figure~\ref{fig:order}. As we expected, the MAE score of \textit{$1^{st}$-order} is the highest, and the MAE score of \textit{$1^{st}$-to-$3^{rd}$-order} is the lowest, which verifies that the introduction of higher-order many-body interactions is effective in improving the accuracy of molecular attribute prediction. Besides, the relative improvement brought by adding the $3^{rd}$-order track is smaller than that by adding the $2^{nd}$-order track, which we speculate is due to the marginal effect.

% \begin{table}
% \caption{MAE of different order of GEM-2 variants on PCQM4Mv2. Lower is better.}
% \centering
% \small
% \label{tab:interaction_order}
% \begin{tabular}{l|ccc}
% \toprule
% Order  & Only $1^{st}$-order & $1^{st}$-to-$2^{nd}$-order & $1^{st}$-to-$3^{rd}$-order \\
% \midrule
% MAE & 0.0930 & 0.0889 & 0.880 \\
% \bottomrule
% \end{tabular}
% \end{table}

\subsubsection{Effects of the Levels of Long-Range Interactions}
\label{sec:long}
We utilize the full-range interaction through the attention mechanism in the above experiments. It would also be interesting to investigate the contributions of different levels of long-range interactions. The level of the long-range interactions is simply controlled by applying attention mechanism to those atoms whose topological distance is within a specific value (level). For example, the level of $k$ means the attention mechanism is only applied between atoms with topological distance less than or equal to $k$. The level of $1$ means the attention mechanism is only applied between atoms connected by chemical bonds. The MAE scores of the GEM-2 variants of different levels of long-range interactions, from level from 1 to 13, are shown in Figure~\ref{fig:hop}. The MAE scores of GEM-2 get lower with the increase of levels, which again demonstrates the crucial of full-range interactions, especially long-range interactions.

% \begin{table}[t]
% \caption{MAE of different $hop\_cut$ of GEM-2 variants on PCQM4Mv2. Lower is better.}
% \centering
% \small
% \label{tab:hop_mask}
% \begin{tabular}{l|ccccccc}
% \toprule
% $hop\_cut$ & 1      & 3       & 5        & 7      & 9  & 11 & 13 \\
% \midrule
% MAE & 0.0883 & 0.0875 & 0.0870 & 0.0871 & 0.0871 & 0.0870 & 0.0872 \\
% \bottomrule
% \end{tabular}
% \end{table}

% \subsubsection{Effects of the Architecture Design of Optimus Block}
% Additional experiments are conducted to analyze the effects of the components in Optimus block. We analyze the impact of the simulated atom coordinates, the for example, how to fuse higher-order messages in Many-body Axial Attention. Please refer to Section~\ref{sec:further_ablation} in Appendix for more details.

\section{Discussion}

\begin{figure}
    \centering
    \includegraphics[width=1.0\linewidth]{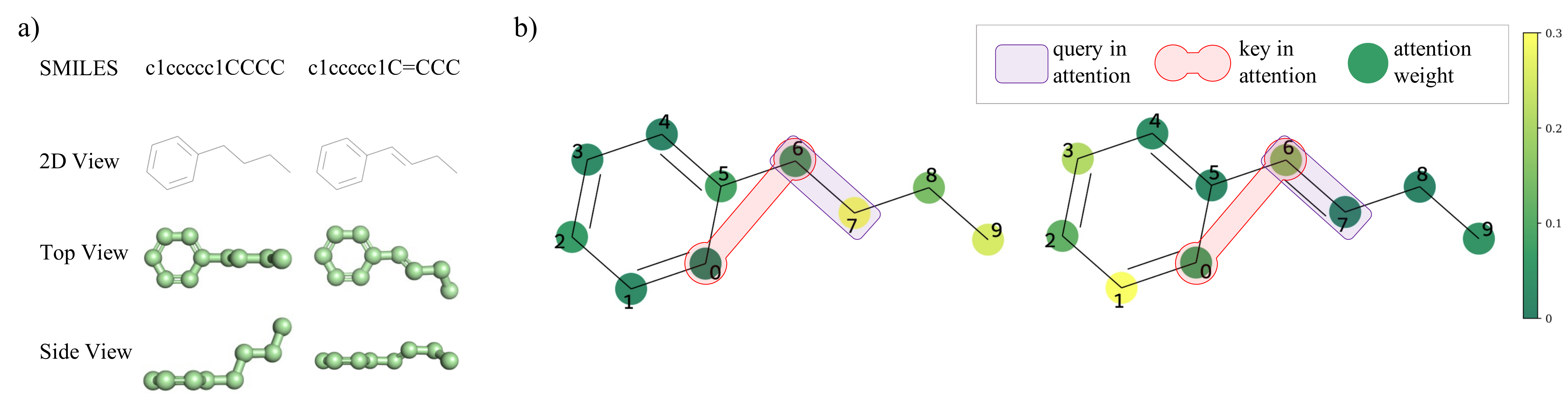}
    \caption{Visualization of attention weights of Many-body Axial Attention of two molecules, n-Butylbenzene (nB) and 1-Phenyl-1-butene (1P1B). a) Smiles, 2D view, top view and side view of the molecules. b) Attention weights of 2-body Axial Attention on $1^{st}$ axis averaged over all heads. Each circle over atom $j$ represents the attention weights between atom pair $(6,7)$ and atom pair $(j,7)$. 
    % e) Attention probabilities of 2-body Axial Attention on 1st axis of all 8 heads. The element in row $head\_id=a$ and column $atom\_id=j$ represents the attention probability between atom pair $(6,7)$ and atom pair $(j,7)$ on $a^{\text{th}}$ head.
}
    \label{fig:attn_weights}
\end{figure}

Our model has achieved SOTA results on multiple tasks, so an interesting question will be what our model has learned, especially in the high-order track, which is the extra part compared with other models. In order to answer this question, it is natural to regard the attention weights in the model as the predictive relevance of certain atoms and bonds. Then, the visualization of these attention weights makes the model explainable to some extent. This idea has been widely used to interpret attention-based models\cite{Jiménez-Luna_Grisoni_Schneider_2020}. 
Here, we designed two comparable molecules. One is the n-Butylbenzene (nB, shown in Figure~\ref{fig:attn_weights}a, left), and another is the 1-Phenyl-1-butene (1P1B, Figure~\ref{fig:attn_weights}a, right), in which the single bond between atom 6 and 7 in nB is modified to a double bond. The 3D structures of the two molecules are generated by the Merck molecular force field (MMFF94)\cite{halgren1996merck} function in the RDKit tool. The difference between the two molecules is that the double bond between atoms 6 and 7 in 1P1B forms a conjugation system with the benzene, making the whole molecular structure closer to the benzene plane (best view in the Side View of Figure~\ref{fig:attn_weights}a). In this comparable case, we expect the attention weights in our model can reflect two important differences between the two molecules: 1) the 3D structure of nB is symmetric while the 1P1B is not, and 2) the attention of the modified double bond in 1P1B should be more on the left benzene instead of the right two carbons. As shown in Figure~\ref{fig:attn_weights}b, we plot the attention weights of 2-body Axial Attention on $1^{st}$ axis given the atom pair $(6,7)$ as the query of attention. As we expected, the attention weights of nB (Figure~\ref{fig:attn_weights}b, left) are highly symmetric, where the attention weights on atoms 2 and 4 are nearly equal with those on atoms 1 and 0, respectively. On the contrary, the attention weights of 1P1B are quite different on these atoms (Figure~\ref{fig:attn_weights}b, right). Moreover, we can see that the highest attention weights within 1P1B are on atoms 1 and 3, or on the benzene, reflecting the high attention of our model on the conjugated effect. To summarise, this visualization demonstrates that the information learned on the many-body track shows good agreement with the well-established chemistry knowledge.

\section{Related Work}
Existing works for molecular modeling can be simply categorized from two prospective: whether considering the many-body interactions and whether considering long-range interactions.

\textit{Interactions between 1-Bodies v.s. Interactions between Many-Bodies.} 
The mainstream works\cite{DBLP:conf/iconip/DanelSTSSSM20:SGCN,doi:10.1021/acs.jmedchem.9b00959:AttentiveFP,DBLP:journals/corr/abs-2002-08264:MAT,DBLP:journals/corr/abs-2108-03348:EGT} focus on the interactions between $1$-bodies (atoms), applying the graph neural networks (GNNs) or transformer-style models to learn the node representations, i.e., the atom representations. Through multiple stacked model blocks, the messages of the atoms are exchanged step by step. A few advanced works \cite{DBLP:conf/iclr/KlicperaGG20:DimeNet,schutt2021equivariant:painn,DBLP:conf/icdm/ShuiK20:HMGNN} attempt to incorporate geometric information, e.g., bond angles, by modeling interactions between atom pairs ($2$-bodies). For example, DimeNet\cite{DBLP:conf/iclr/KlicperaGG20:DimeNet} propose a message passing scheme that uses the directional information by transforming messages based on the angle between them. In particular, our previous work, GEM \cite{fang2022geometry:GEM}, has demonstrated the positive effect of modeling many-body interaction for general property prediction by constructing two graphs: an atom-bond graph and a bond-angle graph.

\textit{Short-Range Interactions v.s. Long-Range Interactions.} 
Most studies \cite{DBLP:conf/iconip/DanelSTSSSM20:SGCN, DBLP:journals/corr/abs-2006-07739:DeeperGCN,brossard2020graph:gine,DBLP:conf/nips/SchuttKFCTM17:SchNet} regard a molecule as a graph by taking the atoms as the nodes of the graph and apply graph neural networks (GNNs) for molecular modeling. They concentrate on the local interactions, i.e., the short-range interactions, between the atoms. For instance, some works\cite{doi:10.1021/acs.jmedchem.9b00959:AttentiveFP,brossard2020graph:gine} regard two atoms are connected if a chemical bond links them. While the studies \cite{DBLP:conf/iclr/KlicperaGG20:DimeNet,DBLP:conf/nips/SchuttKFCTM17:SchNet} connect two atoms if they are spatially close. Even though partial long-range interactions between the atoms can be inferred through the step-by-step message-passing technique, some information may be lost during the multi-step transition. Such an indirect manner could harm the modeling of the interactions between the indirectly connected atoms, especially those far away. To address this issue, exploiting the Transformer-style architectures \cite{DBLP:journals/corr/abs-2002-08264:MAT,DBLP:journals/corr/abs-2012-09699:GT,DBLP:journals/corr/abs-2108-03348:EGT,DBLP:journals/corr/abs-2205-12454:GPS} to directly capture the full-range interaction has attracted increasing attention. For example, EGT\cite{DBLP:journals/corr/abs-2108-03348:EGT} \cite{DBLP:journals/corr/abs-2107-08773:CoMPT} applied the global self attention directly on molecular graphs with a generalized positional node encoding scheme. While CoMPT\cite{DBLP:journals/corr/abs-2107-08773:CoMPT} further reinforced the message interactions between nodes and edges by adding a node-edge interaction module on the attention mechanism.

\section{Conclusion}
Molecular property prediction serves as a fundamental task in drug and material industries, which in essence, can be determined by its own electronic structure and described by the Schrödinger equation. Inspired by the calculation methods of Schrödinger equation, we clarify the importance of full-range many-body interactions for molecular property prediction and propose deep learning methods to comprehensively model such complex interactions for the improvement of molecular modeling. We propose a novel network architecture called GEM-2. The main component of GEM-2, i.e., Optimus block, simultaneously exploits multiple tracks to update the representations of the many-bodies of different orders, with each track learning the interactions of the many-bodies of a specific order. Many-body Axial Attention is designed to reduce the computational cost of the full-range interactions without loss of effectiveness. Extensive experiments on two kinds of benchmarks exhibit that GEM-2 significantly outperforms the mainstream baselines, demonstrating the effectiveness of full-range many-body interaction modeling for molecular property prediction.

\section*{Code Availability}
The source code of this study is available at GitHub repository (\url{https://github.com/PaddlePaddle/PaddleHelix/tree/dev/apps/pretrained\_compound/ChemRL/GEM-2}) to allow replication of the results.

\section*{Data Availability}
For Quantum Chemistry Benchmark, PCQM4Mv2 dataset is publicly available on OGB's official website(\url{https://ogb.stanford.edu/docs/lsc/pcqm4mv2/}). For Drug Discovery Benchmark, LIT-PCBA dataset can be downloaded from\cite{cai2022fp}'s Github repository(\url{https://github.com/idrugLab/FP-GNN/blob/main/Data.rar}). 

\clearpage

\bibliographystyle{plain}
\bibliography{references}  %%% Uncomment this line and comment out the ``thebibliography'' section below to use the external .bib file (using bibtex) .

\clearpage

\appendix

\section{Methodology Details}

\subsection{Detailed Architecture of GEM-2}
\label{sec:overall_alogrithm}
This section lists the detailed model architecture of GEM-2. 

Alg.~\ref{alg:gem-2} shows the overall algorithm of GEM-2, taking $\set{\mathcal{V}^{(m)}}$ as input to predict the property $y$. Here, $p_m$ is the dropout rate.

\begin{algorithm}
\caption{GEM-2}
\label{alg:gem-2}
\begin{algorithmic}[1]
    \Require GEM\_2($\set{\mathcal{V}^{(m)}}$):
        \Comment{$m \in \set{1,...,M}$}
    \State $\bold{X}^{(m)}$ = Dropout$_{p_m}($Embed($\mathcal{V}^{(m)}$))
    \State $\bold{Z}^{(m)}$ = LayerNorm($\bold{X}^{(m)}$)
    \For{$l \in \set{1,...,L}$}
        \State $\set{\bold{Z}^{(m)}}$ = OptimusBlock($\set{\bold{Z}^{(m)}}$)
    \EndFor
    \LineComment{make prediction}
    \State $y$ = MLP(Pool($\bold{Z}^{(1)}$))
    \State \return $y$
\end{algorithmic}
\end{algorithm}

Alg.~\ref{alg:optimus_block} shows the details of a Optimus block, consisting of $M$ tracks. Each track updates the representation of the many-bodies of the corresponding order.
\begin{algorithm}
\caption{Optimus block}
\label{alg:optimus_block}
\begin{algorithmic}[1]
    \Require OptimusBlock($\set{\bold{Z}^{(m)}}$):
        \Comment{$m \in \set{1,...,M}$}
    \State $\hat{\bold{Z}}^{(0)}$ = 0
    \For{$m \in \set{1,...,M}$}
        \LineComment{$m^{th}$-order track}
        \State $\hat{\bold{Z}}^{(m)}$ = OptimusTrack$_{m}$($\hat{\bold{Z}}^{(m-1)}$, $\bold{Z}^{(m)}$, $\bold{Z}^{(m+1)}$)
    \EndFor
    \State \return $\set{\hat{\bold{Z}}^{(m)}}$
\end{algorithmic}
\end{algorithm}

Alg.~\ref{alg:optimus_track} shows the details of $m^{th}$-order track, which consists of a Low2High module, $m$ Many-body Axial Attention modules, and a Feed Forward module.
\begin{algorithm}
\caption{$m^{th}$-order track}
\label{alg:optimus_track}
\begin{algorithmic}[1]
    \Require OptimusTrack$_{m}$($\bold{Z}^{(m-1)}$, $\bold{Z}^{(m)}$, $\bold{Z}^{(m+1)}$):
    % \If{$m$ < 2}
    %     \State $\hat{\bold{Z}}$ = $\bold{Z}^{(m)}$ + Dropout$_{p_m}$(OuterProduct($\bold{Z}^{(m-1)}$))
    % \Else{}
    %     \State $\hat{\bold{Z}}$ = $\bold{Z}^{(m)}$ + Dropout$_{p_m}$(ElementwiseAdd($\bold{Z}^{(m-1)}$))
    % \EndIf
    \State $\hat{\bold{Z}}$ = $\bold{Z}^{(m)}$ + Dropout$_{p_m}$(Low2High($\bold{Z}^{(m-1)}$))
    \For{$a \in \set{1,...,m}$}
        \State $\hat{\bold{Z}}$ += Dropout$_{p_m}$(AxialAttention$_{k}$($\hat{\bold{Z}}$, $\bold{Z}^{(m+1)}$))
    \EndFor
    
    \State $\hat{\bold{Z}}$ += Dropout$_{p_m}$(FeedForward($\hat{\bold{Z}}$))
    \State \return $\hat{\bold{Z}}$
\end{algorithmic}
\end{algorithm}

Alg. \ref{alg:axial_attention} shows the details of Many-body Axial on $k$-th axis. This is a revised version of original attention module, which we apply self-attention along the $k$-th axis of $\bold{Z}^{(m)}$ and efficiently leverage messages from the higher-order $\bold{Z}^{(m+1)}$.

\begin{algorithm}
\caption{Many-body Axial Attention}
\label{alg:axial_attention}
\begin{algorithmic}[1]
    \Require AxialAttention$_{k}$($\bold{Z}^{(m)}$, $\bold{Z}^{(m+1)}$, $c_m=32$, $N_{\text{head}}^{(m)}=8$):
    
    \LineComment{Input Projections}
    \State $\bold{Z}^{(m)}$ = LayerNorm($\bold{Z}^{(m)}$)
    \State $\bold{Z}^{(m+1)}$ = LayerNorm($\bold{Z}^{(m+1)}$)
    
    \State $\bold{q}_i^h$ = Linear($\bold{z}^{(m)}_{i_1,\cdots,i_{k-1},i,i_{k+1},\cdots,,i_m}$)
        \Comment{$\textbf{q}_i^h \in \mathbb{R}^{c_m}$, $h \in \{1,...,N_{\text{head}}^{(m)}\}$}
    \State $\bold{k}_{ij}^h$ = Linear($\bold{z}^{(m)}_{i_1,\cdots,,i_{k-1}ji_{k+1},\cdots,,i_m}$) + Linear($\bold{z}^{(m+1)}_{i_1,\cdots,i_{k-1},i,j,i_{k+1},\cdots,i_m}$)
        \Comment{$\textbf{k}_{ij}^h \in \mathbb{R}^{c_m}$}
    \State $\bold{v}_{ij}^h$ = Linear($\bold{z}^{(m)}_{i_1,\cdots,i_{k-1},j,i_{k+1},\cdots,i_m}$) + Linear($\bold{z}^{(m+1)}_{i_1,\cdots,i_{k-1},i,j,i_{k+1},\cdots,i_m}$)
        \Comment{$\textbf{v}_{ij}^h \in \mathbb{R}^{c_m}$}
    
    \LineComment{Attention}
    \State $\alpha_{ij}^h$ = softmax($\bold{q}_i^T \bold{k}_{ij})$
    \State $\bold{o}_i^h$ = $\sum_{j}{\alpha_{ij} \textbf{v}_{ij}^h}$
    
    \LineComment{Output projection}
    \State $\hat{\bold{z}}_{i_1,\cdots,i_{k-1},i,i_{k+1},\cdots,i_m}$ = Linear(concat$_{h}$($\bold{o}_i^h$))
        \Comment{$\bold{z}_{i_1,\cdots,i_{k-1},i,i_{k+1},\cdots,i_m} \in \mathbb{R}^{c_m\times N_{\text{head}}^{(m)}}$}
    \State \return $\hat{\bold{Z}}$
\end{algorithmic}
\end{algorithm}

Alg. \ref{alg:low2high} shows the details of the Low2High module. When $m=1$, there's no need for a Low2High module. When $m=2$, we utilize an OuterProduct as the Low2High module, which is computationally expensive but has relatively high capacity. When $m>2$, we utilize an ElementwiseAdd as the Low2High module with low computational cost.
\begin{algorithm}
\caption{Low2High module}
\label{alg:low2high}
\begin{algorithmic}[1]
    \Require Low2High($\bold{Z}^{(m-1)}$,$c_m$):
    
    \If{$m$ == 1}
        \State $\bold{Z}^{(m)}$ = 0
    \ElsIf{$m$ == 2}
        \State $\bold{Z}^{(m)}$ = OuterProduct($\bold{Z}^{(m-1)}$)
    \Else{}
        \State $\bold{Z}^{(m)}$ = ElementwiseAdd($\bold{Z}^{(m-1)}$)
    \EndIf

    \State \return $\bold{Z}^{(m)}$
\end{algorithmic}
\end{algorithm}

\begin{algorithm}[h!]
\caption{Outer product}
\label{alg:out_product}
\begin{algorithmic}[1]
    \Require OuterProduct($\bold{Z}^{(1)}$,$c_{\text{outer}}$,$c_m$):
    
    \State $\bold{z}_i^{(1)} = \text{LayerNorm}(\bold{z}_i^{(1)})$
    \State $\bold{z}_{ij}^{(2)} = \text{flatten}(\text{Linear}(\bold{z}_i^{(1)}) \otimes \text{Linear}(\bold{z}_j^{(1)}))$
        \Comment{$\bold{z}_{ij}^{(2)} \in \mathbb{R}^{c_{\text{outer}} \cdot c_{\text{outer}}}$}
    
    \State $\bold{z}_{ij}^{(2)} = \text{Linear}(\bold{z}_{ij}^{(2)})$
        \Comment{$\bold{z}_{ij}^{(2)} \in \mathbb{R}^{c_m}$}

    \State \return $\bold{Z}^{(2)}$
\end{algorithmic}
\end{algorithm}

\begin{algorithm}[h!]
\caption{Element-wise Addition}
\label{alg:ElementwiseAdd}
\begin{algorithmic}[1]
    \Require ElementwiseAdd($\bold{Z}^{(m-1)}$,$c_m$):
    
    \State $\bold{z}_{i_1,\cdots,i_{m-1}}^{(m-1)} = \text{LayerNorm}(\bold{z}_{i_1,\cdots,i_{m-1}}^{(m)})$
    \State $\bold{z}_{i_1,\cdots,i_m}^{(m)}$ = Linear($\bold{z}_{i_2,\cdots,i_m}^{(m-1)}$) + Linear($\bold{z}_{i_1,i_3,\cdots,i_m}^{(m-1)}$) + $\cdots$ + Linear($\bold{z}_{i_1,\cdots,i_{m-1}}^{(m-1)}$)
        \Comment{$\bold{z}_{i_1,\cdots,i_m}^{(m)} \in \mathbb{R}^{c_m}$}
    
    \State $\bold{z}_{i_1,\cdots,i_m}^{(m)}$ = Linear(Activation($\bold{z}_{i_1,\cdots,i_m}^{(m)}$))
        \Comment{$\bold{z}_{i_1,\cdots,i_m}^{(m)} \in \mathbb{R}^{c_m}$}

    \State \return $\bold{Z}^{(m)}$
\end{algorithmic}
\end{algorithm}

\subsection{Input Features}
\label{sec:input_features}
All the features are extracted by RDKit. The feature list is shown in Table~\ref{tab:features}. There are two types of features: discrete features and continuous features. For discrete features, we use the one-hot operation to embed the features into one-hot vectors. While for the continuous features, we use the Radial Basis Function\cite{buhmann2003radial} to expand each continuous value $x$ into a vector $\bold{e} \in \mathbb{R}^M$:
\begin{equation}
    e_m(x) = \exp(-\gamma||x - \mu_m||^2),
\label{eq:rbf}
\end{equation}
where $\gamma$ controls the shape of the radial kernel, and we set $\gamma=10$. $\{\mu_m\}$ is a list of centers ranging from the minimum value to the maximum value of corresponding features with stride of 0.1.

\begin{table}
    \caption{Input features of GEM-2.}
    \small
    \label{tab:features}
    \centering
    \setlength\tabcolsep{2pt}{
    \begin{tabular}{c|ccc}
        \toprule
        \multicolumn{1}{c|}{} & Feature & Description & Shape \\
        \midrule
        
        \multirow{7}{*}{$1$-body} & atom type & type of atom (e.g., C, N, O), by atomic number (one-hot) & $(N,119)$ \\
          &  aromaticity & whether the atom is part of an aromatic system (one-hot) & $(N,2)$ \\
         & formal charge & electrical charge (one-hot) & $(N,16)$ \\
         & chirality tag & CW, CCW, unspecified or other (ont-hot) & $(N,4)$ \\
        &  degree & number of covalent bonds (one-hot) & $(N,11)$ \\
         &  number of hydrogens & number of bonded hydrogen atoms (one-hot) & $(N,9)$ \\
        &  hybridization & sp, sp$^{2}$, sp$^{3}$, sp$^{3}$d or sp$^{3}$d$^{2}$ (one-hot) & $(N,5)$ \\
        \midrule
        
        \multirow{5}{*}{$2$-body} & bond dir & begin ash, begin wedge, etc. (one-hot) & $(N,N,7)$ \\
        &  bond type & single, double, triple or aromatic (one-hot) & $(N,N,4)$ \\
        & in ring & whether the bond is part of a ring (one-hot) & $(N,N,2)$ \\
        & pair hop distance & hop distance between the atom pair (float) & $(N,N,1)$ \\
        & pair distance & simulated  geometric distance between the atom pair (float) & $(N,N,1)$ \\
        \midrule
        
       \multirow{2}{*}{$3$-body} & triplet angle &  three angles of the triangle formed by triplet (float) &  $(N,N,N,3)$ \\
        & triplet hop distance & three hop distances of atom $i$ to $j$, $i$ to $k$ and $j$ to $k$ (float) & $(N,N,N,3)$ \\
        \bottomrule
    \end{tabular}}
\end{table}

\subsection{Memory and Computation Cost}
\label{sec:memory_computation_cost}
%Although the Many-body Axial Attention is much more efficient than the original full attention on the many-bodies, it's still computational heavier than methods focusing on atom interactions. 
We take the $m$-body Axial Attention on $k^{th}$-axis as an example to analyze the memory and computational cost. For memory consumption, it requires $\mathcal{O}(\max{(N^mc_m,N^{m+1}N_{\text{head}}^{(m)}}))$ to store the queries, keys, values, and the attention weights of the attention mechanism, where $N_{\text{head}}^{(m)}$ is the number of heads in the attention. For the computational cost, it requires the maximal computation of $\mathcal{O}(\max{(N^mc_m^2,N^{m+1}c_m}))$ in total, including the calculation of the projection of the queries, keys, and values, as well as the matrix multiplication of the keys and values. To train with large batches and save memory consumption, we utilize the recompute strategy in PaddlePaddle\footnote{\url{https://www.paddlepaddle.org.cn/en}}, which keeps only the inputs and outputs of each Optimus block during the feed-forward of the network, and every intermediate variable within the Optimus block is removed and recomputed during gradient back-propagation.

\section{Experimental Details}

\subsection{LIT-PCBA}

The description of the targets in LIT-PCBA in shown in Table~\ref{tab:pcba}.

\begin{table}
\centering
\caption{Statistics of LIT-PCBA dataset.}
\label{tab:pcba}
\begin{tabular}{llcc}
\toprule
         & Target                                       & \#Actives & \#Inactives \\
\midrule
ADRB2       & Beta2 adrenergic receptor                    & 17      & 312,483    \\
ALDH1       & Aldehyde dihydrogenase 1                     & 7,168    & 137,965    \\
ESR\_ago    & Estrogen receptor $\alpha$                         & 13      & 5,583      \\
ESR\_antago & Estrogen receptor $\alpha$                         & 102     & 4,948      \\
FEN1        & FLAP Endonuclease                            & 369     & 355,402    \\
GBA         & Glucocerebrosidrase                          & 166     & 296,052    \\
IDH1        & Isocitrate dihydrogenase                     & 39      & 362,049    \\
KAT2A       & Histone acetyltransferase KAT2A              & 194     & 348,548    \\
MAPK1       & Mitogen-activated protein kinase 1           & 308     & 62,629     \\
MTORC1      & Mechanistic target of rapamycin              & 97      & 32,972     \\
OPRK1       & Kappa opioid receptor                        & 24      & 269,816    \\
PKM2        & Pyruvate kinase muscle isoform 2             & 546     & 245,523    \\
PPARG       & Peroxisome proliferator-activated receptor $\gamma$ & 27      & 5,211      \\
TP53        & Cellular tumor antigen p53                   & 79      & 4,168      \\
VDR         & Vitamin D receptor                           & 884     & 355,388   \\
\bottomrule
\end{tabular}
\end{table}

\subsection{Hyper-parameters}
\label{sec:hyper_parameters}
For model optimization of all our experiments, we utilize the Adam optimizer with a learning rate warm-up and a learning rate decay: the warm-up starts with $1\%$ of the learning rate and linearly scales to $100\%$ for 10 epochs, then the learning rate is kept for 40 epochs. Finally, we continuously decay the learning rate with $50\%$ for every 10 epochs. We train our model for 100 epochs in total. Other settings of hyper-parameters are slightly tuned to gain the best performance for different benchmarks, including learning rate and hidden sizes of GEM-2, as shown in Table \ref{tab:hyper_parameters}. While for the ablation studies on PCQM4Mv2, we fix the hyper-parameters and use a smaller hidden size to save computational consumption. In Section~\ref{sec:many_vs_long}, we set $c_1 = c_2 = c_3 = 128$ and learning rate as $8\cdot 10^{-4}$, while in Section~\ref{sec:order} and \ref{sec:long}, we set $c_1 = c_2 = c_3 = 64$, learning rate as $8\cdot 10^{-4}$ and train for 50 epochs.

\begin{table}
\centering
\caption{Hyper-parameters of GEM-2 for training.}
\label{tab:hyper_parameters}
\begin{tabular}{l|cc}
\toprule
parameter name & PCQM4Mv2 & LIT-PCBA \\
\midrule
batch size               & 512 & 256 \\
learning rate               & $4\cdot 10^{-4}$ & $2\cdot 10^{-4}$ \\
$p_m$     & 0.05 & 0.2    \\
$c_m$     & 256 & 128    \\
\bottomrule
\end{tabular}
\end{table}

\end{document}